\RequirePackage{fix-cm}
\documentclass[twocolumn]{svjour3}          
\smartqed  
\usepackage{graphicx}
\usepackage{color}
\usepackage{amsmath,amsfonts}
\usepackage{algorithmic}
\usepackage{algorithm}
\usepackage{array}
\usepackage{textcomp}
\usepackage{stfloats}
\usepackage{url}
\usepackage{verbatim}
\usepackage{graphicx}
\usepackage{cite}
\usepackage{adjustbox}
\usepackage{booktabs}
\usepackage{subfigure}
\usepackage{multirow}
\usepackage{hyperref}

\journalname{CGI2024} 
\begin{document}

\title{MFDNet: Multi-Frequency Deflare Network for Efficient Nighttime Flare Removal}
\subtitle{}

\author{Yiguo Jiang \and Xuhang Chen \and Chi-Man Pun \and Shuqiang Wang \and Wei Feng}
\institute{
Corresponding author: Chi-Man Pun \\ E-mail: cmpun@umac.mo \\ This work was supported in part by the Science and Technology Development Fund, Macau SAR,
under Grants 0141/2023/RIA2 and 0193/2023/RIA3.}
\date{ }

\maketitle

\begin{abstract}

When light is scattered or reflected accidentally in the lens, flare artifacts may appear in the captured photos, affecting the photos' visual quality. The main challenge in flare removal is to eliminate various flare artifacts while preserving the original content of the image. To address this challenge, we propose a lightweight Multi-Frequency Deflare Network (MFDNet) based on the Laplacian Pyramid. Our network decomposes the flare-corrupted image into low and high-frequency bands, effectively separating the illumination and content information in the image. The low-frequency part typically contains illumination information, while the high-frequency part contains detailed content information. So our MFDNet consists of two main modules: the Low-Frequency Flare Perception Module (LFFPM) to remove flare in the low-frequency part and the Hierarchical Fusion Reconstruction Module (HFRM) to reconstruct the flare-free image. Specifically, to perceive flare from a global perspective while retaining detailed information for image restoration, LFFPM utilizes Transformer to extract global information while utilizing a convolutional neural network to capture detailed local features. Then HFRM gradually fuses the outputs of LFFPM with the high-frequency component of the image through feature aggregation. Moreover, our MFDNet can reduce the computational cost by processing in multiple frequency bands instead of directly removing the flare on the input image. Experimental results demonstrate that our approach outperforms state-of-the-art methods in removing nighttime flare on real-world and synthetic images from the Flare7K dataset. Furthermore, the computational complexity of our model is remarkably low.
\end{abstract}

\section{Introduction}


Photographs taken in nighttime scenes with bright light sources often exhibit flare artifacts, which occur as a result of undesired scattering and reflection of intense light within the camera lens. Light scattering and reflection are common in real camera lenses, particularly in consumer-grade mobile phone cameras. Daily wear and tear, along with the presence of fingerprints and dust, can unintentionally cause light scattering or reflections in the lens. These flare artifacts not only impact the aesthetics of the photograph but also degrade the detailed visual information, hindering image comprehension. As a result, there is a strong demand for a reliable and effective nighttime flare removal algorithm.


The flare patterns in photographs are affected by the lens properties and shooting environment, which encompass factors such as the design of the optics lens, manufacturing imperfections, lens smudges, and the light source's position and angle relative to the lens. The diversity of these factors results in flares with different shapes, positions, and colors. Typical flare artifacts include glare, streaks, bright colored lines, shimmer, saturated blobs, and many others. The variety in appearance of flare artifacts makes it challenging to remove them entirely from a photograph while preserving other content information, especially when multiple flare patterns exist within a single image.

\begin{figure*}[tb]
	\centering
	\begin{minipage}[b]{1.0\textwidth}
	\subfigure[Flare7K real-world dataset]{
		\begin{minipage}[b]{0.48\textwidth}
			\includegraphics[width=1\textwidth]{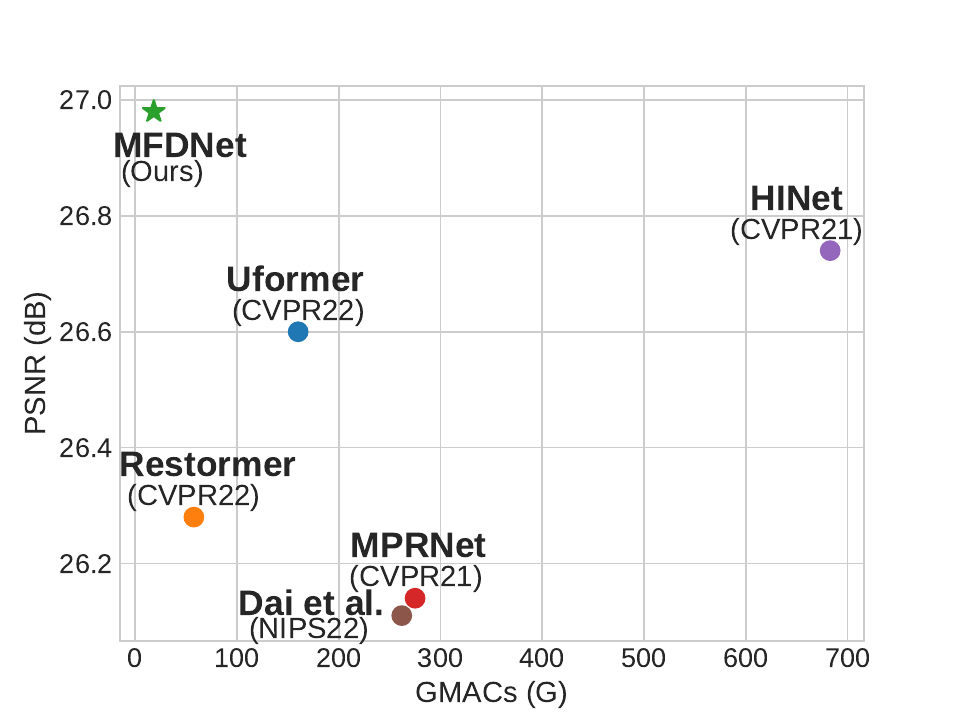}\\ \vspace{-4mm}
		\end{minipage}
	}\hspace{-2mm}
	\subfigure[Flare7K synthetic dataset]{
		\begin{minipage}[b]{0.48\textwidth}
   	 	    \includegraphics[width=1\textwidth]{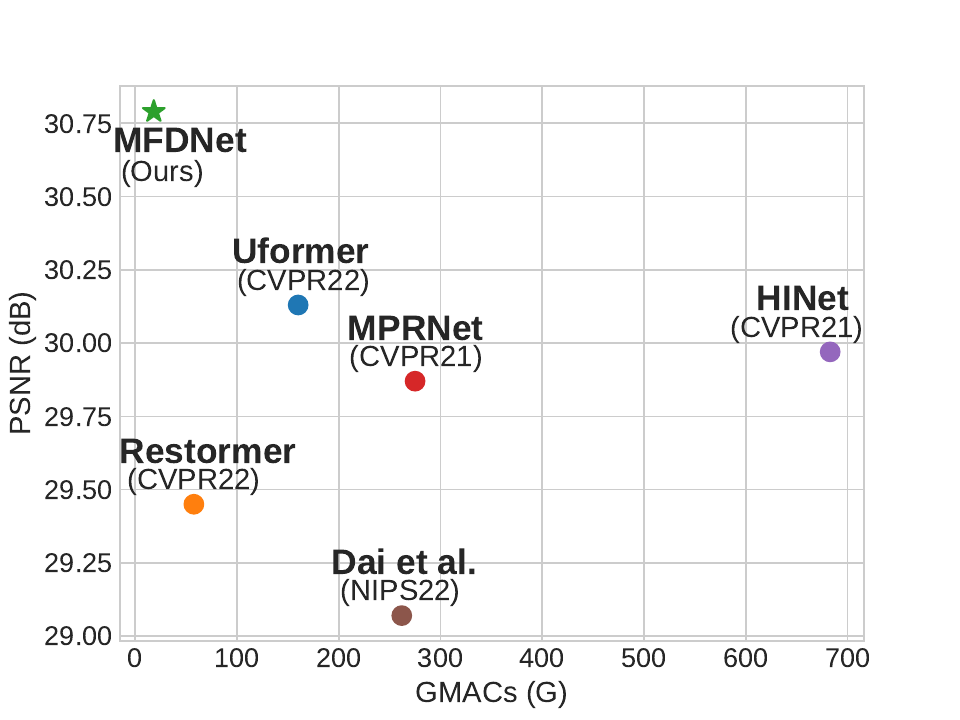}\\ \vspace{-4mm}
		\end{minipage}
		}\hspace{-2mm}
	\end{minipage}\vspace{-4mm}
	\caption{Our MFDNet achieves the state-of-the-art performance
on nighttime flare removal task while being computationally efficient.} \vspace{-4mm}
	\label{fig:teaser}
\end{figure*}


Traditional flare removal methods include hardware-based methods \cite{boynton2003liquid,macleod2010thin} and software-based methods \cite{faulkner1989veiling,seibert1985removal,zhang2018single,asha2019auto,vitoria2019automatic,Chabert2015AutomatedLF}. 
Advanced materials and refined optical designs can contribute to more specialized lenses that reduce flare artifacts.
Applying an anti-reflective coating is another widely used way of reducing flare impact.
Nevertheless, these hardware methods can alleviate some flare effects but cannot wholly remove various flares. Furthermore, these hardware-based methods are not useful in images containing flares and are relatively expensive. 
To address these above problems, some software-based algorithms have emerged for removing flares. These methods usually follow two steps: detecting flares based on their characteristics and then removing them. However, software-based methods struggle to handle a broad range of flare artifacts.

Recently, Deep learning-based methods \cite{wu2021train,dai2022flare7k,chen2021hinet,zamir2021multi,zamir2022restormer,wang2022uformer} for removing flare have emerged. Nevertheless, most of these existing methods view flare removal as a general image restoration task and do not consider effectively decoupling the image's illumination and content information. As a result, these methods might not fully eliminate the flare artifacts or cause degradation of the image content during the removal process. Moreover, these methods directly utilize deep networks to globally manipulate the flare-corrupted image, which results in high computational costs. Consequently, their computational complexity exponentially increases with image resolution, making it unfeasible to apply to high-resolution images, reducing the algorithm's applicability.

In this paper, we propose a lightweight Multi-Frequency Deflare Network (MFDNet) based on the Laplacian Pyramid \cite{burt1987laplacian} for nighttime flare removal. Our proposed method aims to effectively eliminate various flare artifacts while preserving the integrity of the original image. 
Inspired by the reversible frequency-band decomposition framework of a Laplacian Pyramid \cite{burt1987laplacian},
our MFDNet decouples illumination and content information by decomposing the image into low and high-frequency bands. It performs flare removal in the low-frequency part of the image, followed by gradual fusion with the high-frequency part to reconstruct the flare-free image. At the same time, because our method performs flare removal in the low-frequency part, where the resolution is low, it can effectively reduce the computational complexity. As shown in Figure \ref{fig:teaser}, our method of first decoupling the flare-corrupted image and then removing the flare can effectively eliminate the flare artifacts while minimizing the computational complexity.


Specifically, we propose the Low-Frequency Flare Perception Module (LFFPM) for flare removal in the low-frequency part. In the task of flare removal, a large receptive field is essential due to the extensive coverage of the flare. Thus, global information is crucial in accurately identifying the flare. Considering the Transformer's proficiency in capturing long-range pixels, the Low-Frequency Flare Perception Module (LFFPM) utilizes the Transformer for global feature extraction and refinement. In order to alleviate the limitation of Transformers in capturing local dependencies and reduce the model's computational complexity, LFFPM uses a convolution-based encoder-decoder structure to enhance local detailed feature representation. In addition, we propose the Hierarchical Fusion Reconstruction Module (HFRM) for an efficient fusion of high-frequency information. In HFRM, features from the high-frequency component and the results of the Low-Frequency Flare Perception Module (LFFPM) are aggregated at each layer to construct the Laplacian Pyramid for the final reconstruction. 

In summary, the contributions of this paper are as follows: 

1. We propose a lightweight and effective Multi-Frequency Deflare Network (MFDNet) that removes nighttime flare artifacts by decoupling the image's illumination and content information into different frequency bands. 

2. We design the Low-Frequency Flare Perception Module (LFFPM) to remove flares in the low-frequency part, which utilizes convolution to capture local features and self-attention to model long-range dependencies.

3. We design the Hierarchical Fusion Reconstruction Module (HFRM), which gradually aggregates features from the high-frequency bands and LFFPM's results to reconstruct the final flare-free image.

4. Extensive experiments demonstrate that our method achieves state-of-the-art performance on nighttime flare removal task while maintaining low computational complexity.

The subsequent sections are structured as follows. Section \ref{related} discusses related works. Section \ref{method} is devoted to the details of our proposed MFDNet elaborately. Section \ref{experiment} presents our extensive experiments and analysis, and the conclusion is summarized in Section \ref{conclusion}.

\section{Related Work} \label{related}

\subsection{Image Restoration}
During the process of capturing photographs, various factors, including unfavorable weather conditions, optics-induced diffraction, and relative motion between the camera and object, can lead to the deterioration of image quality. This degradation results in the loss of important information in the captured image, necessitating the restoration of the image to its original quality. Typical image restoration tasks include but are not limited to image deblurring \cite{chen2023deep}, image denoising \cite{zhang2023compressive}, image dehazing \cite{chougule2023agd}, 
rain removal \cite{ragini2023detformer}, reflection removal \cite{9381898}, shadow removal \cite{9408597}, and more. Recently, some state-of-the-art image restoration methods \cite{zamir2022restormer,zamir2021multi,wang2022uformer,chen2021hinet,chen2021pre,liang2021swinir,Li2021OnET,tu2022maxim,8485303,9173764,luo2023hir} have appeared to handle different image restoration tasks.

\subsection{Flare Removal}

\subsubsection{Traditional methods}

Traditional flare removal methods include hardware-based methods and software-based methods. Most hardware-based methods aim to reduce flare artifacts by improving optical designs and camera lens materials. Boynton et al. \cite{boynton2003liquid} propose a fluid-filled camera lens to alleviate flare artifacts caused by light reflections. Macleod et al. \cite{macleod2010thin} employ a neutral density filter to minimize reflective flare artifacts. Another commonly used hardware-based approach is to apply an anti-reflective coating to the camera lens. However, this coating may interfere with other coatings like anti-scratch and anti-fingerprint, and it is usually only designed to work for specific light wavelengths and angles of incidence. While these specific hardware approaches can eliminate some lens flare artifacts, they can often not address unforeseeable flares such as those generated by fingerprints or dust on the lens. In addition, these hardware-based methods are usually expensive, and none of them can deal with photographs that already exhibit flare artifacts.


In response to the above problems, some algorithms have been proposed to remove flare artifacts. Seibert et al. \cite{seibert1985removal} and Faulkner et al. \cite{faulkner1989veiling} propose to use deconvolution to remove flare artifacts. Zhang et al. \cite{zhang2018single} propose a method to remove flare by separating the image into a flared part and a scene part.
Other methods \cite{asha2019auto,vitoria2019automatic,Chabert2015AutomatedLF} employ a two-step approach, where they detect flare based on their features and subsequently eliminate flare artifacts while reconstructing affected areas through inpainting \cite{criminisi2004region}. However, these approaches might wrongly identify bright regions as flare artifacts. Additionally, they may not be effective in dealing with various patterns of flare artifacts in complex scenarios.

\begin{figure*}[!htb]
\centerline{\includegraphics[width=1 \linewidth]{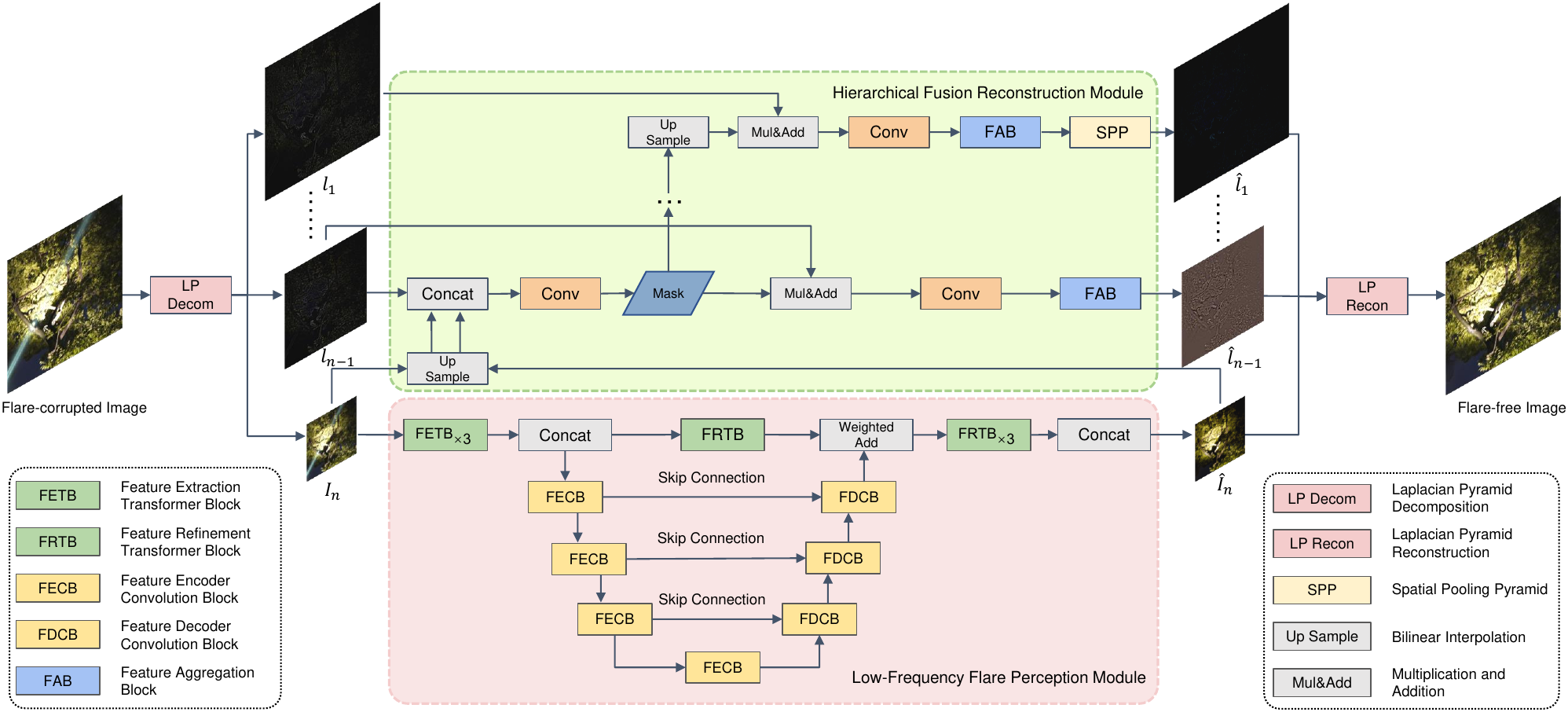}}
\caption{The structure of our proposed MFDNet. Given a nighttime flare-corrupted image, our MFDNet first decouples its content and
illumination information by decomposing this image into high and
low-frequency bands. Then MFDNet performs flare removal in the low-
frequency part of the image, followed by gradual fusion with
the detailed high-frequency part to reconstruct the final flare-free image.} \label{MF}
\end{figure*}

\subsubsection{Deep learning-based methods}

Recently, deep learning-based methods have achieved good results in some low-level image restoration tasks \cite{zamir2022restormer,zamir2021multi,wang2022uformer,chen2021hinet,chen2021pre,liang2021swinir,Li2021OnET,tu2022maxim,8485303,9173764,10208804}. The success of these deep learning methods typically depends on domain-specific image training datasets. However, collecting large-scale pairs of flare-corrupted and flare-free images from real-world scenes can be a labor-intensive and time-consuming process. Consequently, the development of deep learning-based flare removal algorithms has been sluggish, with only a limited amount of related work emerging in recent years. 


Wu et al. \cite{wu2021train} proposed the first deep learning-based method for daytime flare removal utilizing a U-Net \cite{ronneberger2015u} model to reconstruct flare-free images. Their approach involved a post-processing step to reintegrate the light source into the restored image. Many subsequent methods, including ours, adopted a similar pipeline by first removing the flare and then blending the light source back into the image. 
Dai et al. \cite{dai2022flare7k} proposed a method for synthesizing flare by simulating the optical principles of nighttime flare generation. This enables them to construct datasets consisting of paired flare-corrupted and flare-free images. They created Flare7K, the first benchmark dataset for nighttime flare removal, which serves as a valuable resource for tackling this challenging task. With the proposed Flare7K dataset, they follow Wu et al. \cite{wu2021train} and train a U-Net \cite{ronneberger2015u} as a baseline network. Meanwhile, they train some state-of-the-art image restoration methods in the Flare7K dataset, including HINet \cite{chen2021hinet}, MPRNet \cite{zamir2021multi}, Restormer \cite{zamir2022restormer}, and Uformer \cite{wang2022uformer} to build the flare removal benchmark. 
FF-Former \cite{10208804} proposes a U-shape network based Fast Fourier Convolution (FFC) for nighttime flare removal, which addresses the
issue of limited receptive field in traditional window-based Transformer approaches.

HINet \cite{chen2021hinet} integrates Instance Normalization (IN) into the basic module to build the HIN module, which improves the performance of the image restoration network. MPRNet \cite{zamir2021multi} proposes a multi-stage architecture that incrementally learns recovery functions for degraded inputs, thereby dividing the entire restoration process into more manageable steps. Restormer \cite{zamir2022restormer} proposes an encoder-decoder Transformer model to learn multi-scale representations of high-resolution images without decomposing them into local windows. Uformer \cite{wang2022uformer} proposes a universal U-shaped Transformer for various image restoration tasks, which is built on the basic locally-enhanced window Transformer module and is efficient and effective.
In Section \ref{experiment}, we compare our MFDNet with these state-of-the-art methods in Flare7K benchmark.



\section{Methodology} \label{method}

\subsection{Overview}

For the nighttime flare removal task, we propose a lightweight Multi-Frequency Deflare Network (MFDNet) based on the Laplacian Pyramid. As shown in Figure \ref{MF}, our MFDNet consists of two primary modules: the Low-Frequency Flare Perception Module (LFFPM) and the Hierarchical Fusion Reconstruction Module (HFRM). Next, we first describe the overall pipeline of MFDNet, and then we detail the LFFPM in Section \ref{sec:lfpm} and the HFRM in Section \ref{sec:hfrm}.

The Laplacian Pyramid (LP) \cite{burt1987laplacian} is a frequency-band image decomposition technique derived from the Gaussian Pyramid (GP). The main idea of the LP method is to decompose an image linearly into high-frequency and low-frequency bands. And based on the LP, the image reconstruction process can be implemented with precision and reversibility. Specifically, the LP is obtained by calculating the difference between adjacent layers in the Gaussian Pyramid. According to \cite{burt1987laplacian,liang2021high}, the lumination information of the image is more related to the low-frequency band and the high-frequency component contains more detailed content information such as textures. Inspired by the above properties of LP, we decouple the image's illumination and content information, remove the flare artifacts in the low-frequency part of the image, and subsequently fuse the low-frequency flare-free image with the detailed high-frequency information to restore the final flare-free image. The entire process is depicted in Figure \ref{MF}.

Specifically, given a nighttime flare-corrupted image $I \in R^{H \times W \times 3}$, MFDNet first decomposes it into a Laplacian Pyramid, generating a set of different frequency parts $L=[l_1,l_2, \cdots,l_{n-1}]$ and the lowest frequency image $I_n$, where $H \times W$ denotes the spatial dimension, and $n$ is the number of LP's decomposed levels. The components of $L$ have progressively reducing resolutions from $H \times W$ to $\frac{H}{2^{n-1}} \times \frac{W}{2^{n-1}}$, and $I_n$ has $\frac{H}{2^{n}} \times \frac{W}{2^{n}}$ pixels. After getting $I_n$, we input it into the Low-Frequency Flare Perception Module (LFFPM) for flare removal and get a low-frequency flare-free image ${\hat{I}}_n$. $I_n$ and ${\hat{I}}_n$ are then provided to the Hierarchical Fusion Reconstruction Module (HFRM), which fuses each layer of $L$ with ${\hat{I}}_n$ incrementally to create $\hat{L} =[\hat{l}_1,\hat{l}_2, \cdots,\hat{l}_{n-1}]$ for reconstruction. $L$ and $\hat{L}$ share a one-to-one mirror relationship, so the final flare-free image $\hat{I}$ can be reconstructed by ${\hat{I}}_n$ and $\hat{L}$. 
Because our method only needs to remove flares in the low-frequency components, it significantly reduces computational complexity.


\begin{figure}[t]
\centerline{\includegraphics[width=0.8 \linewidth]{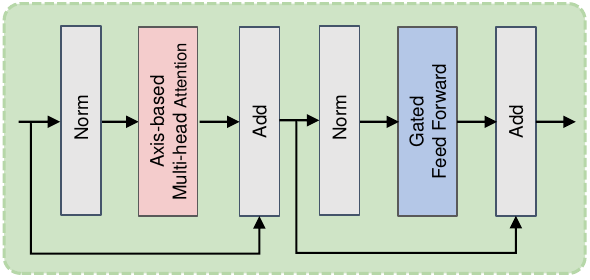}}
\caption{The structure of Feature Extraction Transformer Block (FETB) and Feature Refinement Transformer Block (FRTB).}\vspace{-4mm}
\label{FETB}
\end{figure}

\begin{figure}[t]
\centerline{\includegraphics[width=\linewidth]{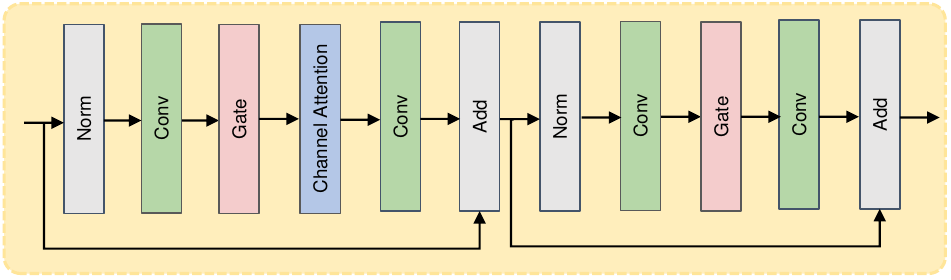}}
\caption{The structure of Feature Encoder Convolution Block (FECB) and Feature Decoder Convolution Block (FDCB).} \vspace{-4mm}
\label{FECB}
\end{figure}

\begin{figure}[t]
\centerline{\includegraphics[width=0.8 \linewidth]{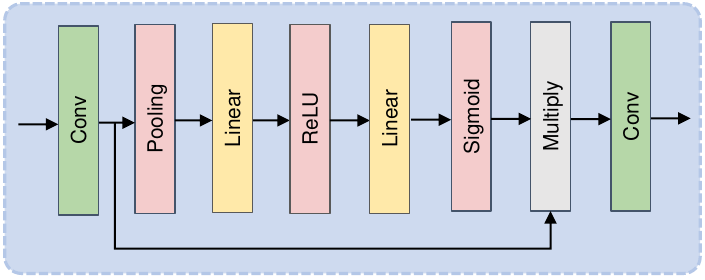}}
\caption{The structure of Feature Aggregation Block (FAB).}\vspace{-4mm}
\label{FAB}
\end{figure}

\subsection{Low-Frequency Flare Perception Module}
\label{sec:lfpm}
The Low-Frequency Flare Perception Module (LFFPM) mainly includes the following parts, the Feature Extraction Transformer Block (FETB) and the Feature Refinement Transformer Block (FRTB) for extracting and refining features, and a U-shaped network with skip connections, composed of the Feature Encoder Convolution Block (FECB) and the Feature Decoder Convolution Block (FDCB).
Transformers have been demonstrated to be more effective than CNNs in modeling long-range dependencies. As depicted in Figure \ref{MF}, we use the Feature Extraction Transformer Block (FETB) in the Low-Frequency Flare Perception Module (LFFPM) to extract global features from the input image $I_n$. Moreover, at the end of LFFPM, we use the Feature Refinement Transformer Block (FRTB) to generate the enhanced features. After multiple FRTBs and FRTBs, we concatenate the output of each block to obtain longer-distance features. Inspired by \cite{wang2023ultra}, FETB and FRTB use axis-based self-attention(ASA) to reduce computational complexity and gated feed-forward network (GFFN) to capture more critical features. Traditional self-attention's computational complexity is quadratic with the input resolution. 
ASA computes self-attention sequentially on the height and width axes across the channel dimension, resulting in linear complexity. GFFN applies GELU and elementwise product to eliminate less relevant features in two parallel paths, then combines the relevant features through element-wise summation. As shown in Figure \ref{FETB}, the computation of FETB and FRTB are represented as:
\begin{equation}
\begin{aligned}
&\mathbf{F}^{\prime}=\mathrm{ASA}(\mathrm{LN}(\mathbf{\mathbf{F}}))+ \mathbf{\mathbf{F}}, \\  
&\mathbf{\hat{F}}=\mathrm{GFFN}(\mathrm{LN}(\mathbf{F}^{\prime}))+\mathbf{F}^{\prime},
\end{aligned}
\end{equation}
where $\mathbf{F}$ is the input of FETB and FRTB. $\mathbf{F}^{\prime}$ and $\mathbf{\hat{F}}$ are the outputs of ASA and GFFN, respectively. $\mathrm{LN}$ represents the layer normalization \cite{ba2016layer}.

In order to alleviate the limitation of Transformers in capturing local dependencies, the Low-Frequency Flare Perception Module (LFFPM) uses a convolution-based U-shape encoder-decoder structure consisting of Feature Encoder Convolution Block (FECB) and Feature Decoder Convolution Block (FDCB) to enhance detailed feature representation, as shown in the Figure \ref{MF}. Additionally, to ensure that both the global and local features are adequately fused, the Low-Frequency Flare Perception Module (LFFPM) automatically learns the weighted summation of the global long-distance features extracted by FETB and the output of FDCB. Inspired by \cite{chen2022simple}, FECB and FDCB are designed as simple and efficient nonlinear activation networks. As shown in Figure \ref{FECB}, FECB and FDCB mainly include the following parts: layer normalization \cite{ba2016layer}, convolution, Gate, and Channel Attention \cite{hu2018squeeze}. Gate divides the feature map into two parts in the channel dimension and multiplies them. The formula of Gate is:
\begin{equation}
\mathrm{Gate}(\mathbf{X},\mathbf{Y})=\mathbf{X} \odot \mathbf{Y},
\end{equation}
where $\mathbf{X}$ and $\mathbf{Y}$ are feature maps.

The Channel Attention(CA) mechanism can capture global information efficiently. The formula of Channel Attention is:
\begin{equation}
\mathrm{CA}(\mathbf{X})=\mathbf{X} \ast W \mathrm{Pool}(\mathbf{X}),
\end{equation}
where $\mathbf{X}$ denotes the feature map, $W$ is fully-connected layers, and $\mathrm{Pool}$ represents the global average pooling. $\ast$ is channel-wise product operation.


\subsection{Hierarchical Fusion Reconstruction Module}
\label{sec:hfrm}
As shown in Figure \ref{MF}, after obtaining the low-frequency flare-free image ${\hat{I}}_n$ through the Low-Frequency Flare Perception Module (LFFPM), the Hierarchical Fusion Reconstruction Module (HFRM) fuses ${\hat{I}}_n$ with the high-frequency parts $L=[l_1,l_2, \cdots,l_{n-1}]$ layer by layer to obtain $\hat{L} =[\hat{l}_1,\hat{l}_2, \cdots,\hat{l}_{n-1}]$ for reconstruction.

We concatenate the upsampled $I_n$ and ${\hat{I}}_n$ with $l_{n-1}$, and then input them into a simple network to learn a mask $M_{n-1}$ to guide the fusion process. The specific fusion process is as follows:
\begin{equation}
{{l}^{\prime}}_{n-1}=l_{n-1} \otimes \mathbf{M}_{n-1} + l_{n-1},
\end{equation}
where $\otimes$ represents the pixel-wise multiplication.
To refine the fused features further, we perform a dilated convolution operation on ${{l}^{\prime}}_{n-1}$, then use the Feature Aggregation Block (FAB) to re-weight the significance of features. 

Through the above steps, the low-frequency flare-free image ${\hat{I}}_n$ achieves the feature fusion operation with one layer in the high-frequency parts $L=[l_1,l_2, \cdots,l_{n-1}]$. By analogy, we need to upsample $\mathbf{M}_{n-1}$ through linear interpolation and perform the same fusion operation with the next layer in the high-frequency parts $L=[l_1,l_2, \cdots,l_{n-1}]$. Behind the Feature Aggregation Block (FAB) in the final layer, we use the Spatial Pooling Pyramid (SPP) \cite{he2015spatial} to facilitate remixing multi-context features. This way, we get $\hat{L} =[\hat{l}_1,\hat{l}_2, \cdots,\hat{l}_{n-1}]$ to reconstruct the final flare-free image.

As shown in Figure \ref{FAB}, the Feature Aggregation Block (FAB) structure is simple. Inspired by \cite{cun2020towards}, we use a squeeze-and-excitation block \cite{hu2018squeeze} in FAB to learn the weights of different channel features through some linear layers and pooling layers, which can automatically preserve essential features. Additionally, convolution is used to squeeze the features and match the original channels.

\subsection{Loss Function}
We train our model with different losses, including the mean square error (MSE) loss $L_M$, structural similarity loss $L_{S}$ \cite{wang2004image}, and perceptual loss \cite{zhang2018unreasonable}.



Given the final output image $I_{out}$ and ground truth image $I_{gt}$, the perceptual loss $L_P$ is defined as:
\begin{equation}
L_P = \sum_l || F_l(I_{out})-F_l(I_{gt}) ||_2^2,
\label{eqn:dist}
\end{equation}
where $F$ is a pre-trained AlexNet \cite{krizhevsky2012imagenet} feature extractor. We compute the $l_2$ distance between $F_l(I_{out})$ and $F_l(I_{gt})$ for layer $l$.

In summary, our total loss function can be expressed as:
\begin{equation}
L_{total} = \lambda_{m} L_{M} + \lambda_{s} L_{S} + \lambda_{p} L_{P},
\end{equation}
where we empirically set $\lambda_{m}=1$, $\lambda_{s}=0.3$ and $\lambda_{p}=0.7$ respectively.

\begin{table*}[ht]
\begin{center}
\caption{Quantitative comparison on Flare7K real-world and synthetic datasets in terms of PSNR, SSIM, LPIPS. The best results are highlighted in bold.}\vspace{-4mm}
\adjustbox{width=\linewidth}{
\begin{tabular}{c c| c c c c c c c c c c }
\toprule[0.15em]
\multicolumn{2}{c|}{\multirow{2}{*}{Dataset$\backslash$Method}} & {\multirow{2}{*}{Input}}& Zhang  & Sharma & Wu   & Dai  & HINet  & MPRNet & Restormer &Uformer & \textbf{MFDNet} \\
 & & &\cite{zhang2020nighttime} & \cite{sharma2021nighttime} & \cite{wu2021train} & \cite{dai2022flare7k} & \cite{chen2021hinet} & \cite{zamir2021multi} &  \cite{zamir2022restormer}	 & \cite{wang2022uformer} &  (Ours)  \\
\midrule[0.15em]
~ & PSNR~$\uparrow$ &  22.56& 21.02 &  20.49 & 24.61  &  26.11  &  26.74  & 26.14  & 26.28  & 26.60 & \textbf{26.98} \\
Real-world & SSIM~$\uparrow$ & 0.857& 0.784 &  0.826 & 0.871  &  0.879  &  0.882  & 0.878  & 0.883  & 0.892 & \textbf{0.895}\\
  & LPIPS~$\downarrow$ & 0.078& 0.174 &  0.112 & 0.060  &  0.055  &  \textbf{0.048}  & 0.050  & 0.054  & 0.051 & 0.051 \\
\midrule[0.1em]
~ & PSNR~$\uparrow$ & 22.77& 21.04 &  20.01 & 27.88  &  29.07  &  29.97  & 29.87  & 29.45  & 30.13 & \textbf{30.79} \\
Synthetic & SSIM~$\uparrow$ & 0.921& 0.841 &  0.865 & 0.952  &  0.958  &  0.959  & 0.959  & 0.950  & 0.965 & \textbf{0.966} \\
  & LPIPS~$\downarrow$ & 0.060& 0.136 &  0.111 & 0.031  &  0.022  &  0.021 & 0.020  & 0.025  & 0.020 & \textbf{0.019}  \\
\bottomrule[0.15em]
\end{tabular}
}\vspace{-4mm}
\label{table:overall}
\end{center}
\end{table*}

\begin{figure*}[!htb]
	\centering
	\begin{minipage}[b]{1.0\textwidth}
	\subfigure[Input]{
		\begin{minipage}[b]{0.134\textwidth}
			\includegraphics[width=1\textwidth]{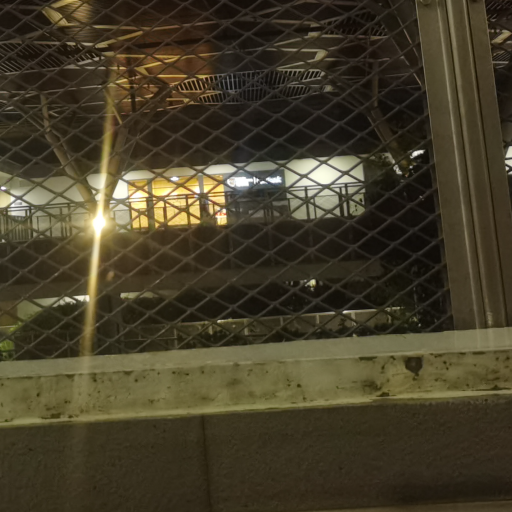}\\
			\includegraphics[width=1\textwidth]{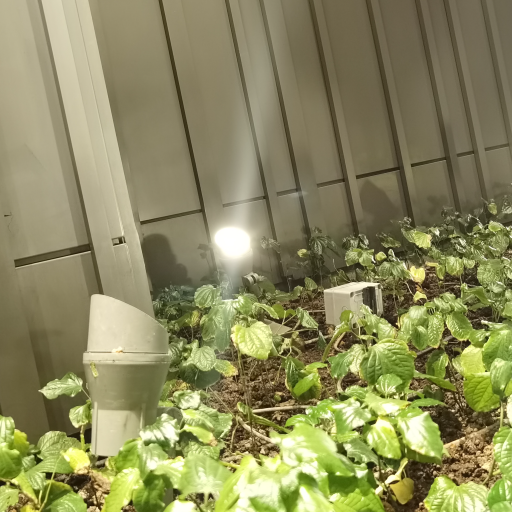}\\ 
			\includegraphics[width=1\textwidth]{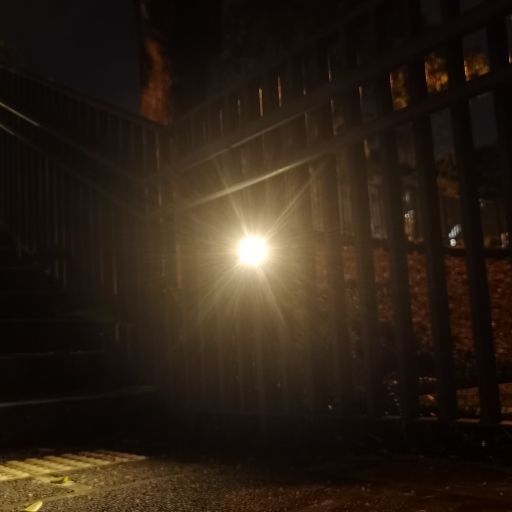}\\ 
			\includegraphics[width=1\textwidth]{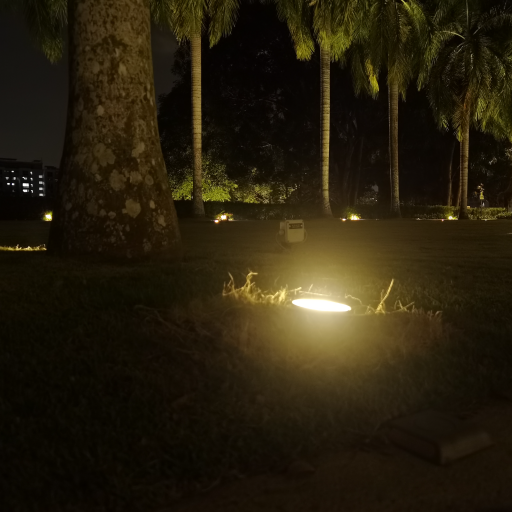}\\
			\includegraphics[width=1\textwidth]{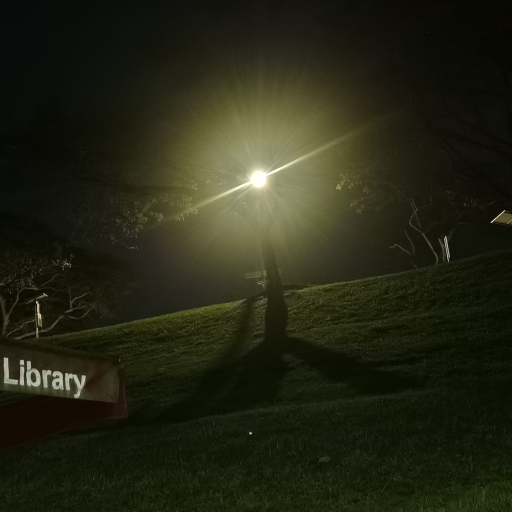}\\ 
			\includegraphics[width=1\textwidth]{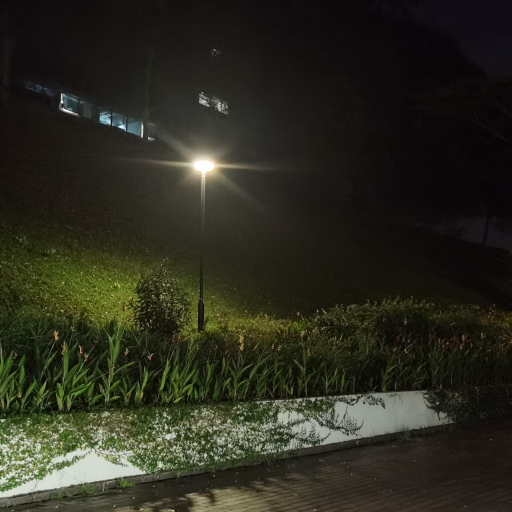}\\ 
			\includegraphics[width=1\textwidth]{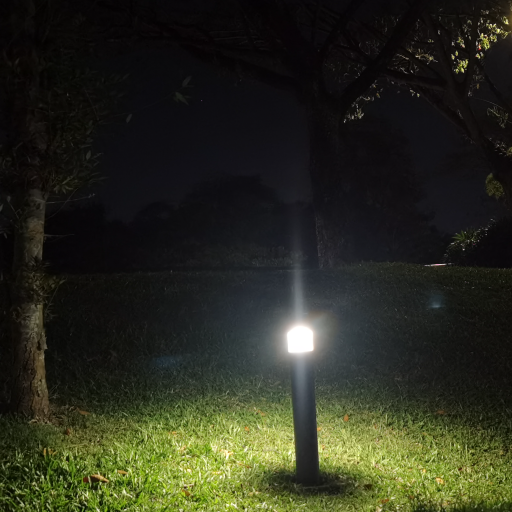}\\ \vspace{-4mm}
		\end{minipage}
	}\hspace{-2mm}
	\subfigure[Sharma\cite{sharma2021nighttime}]{
		\begin{minipage}[b]{0.134\textwidth}
		    \includegraphics[width=1\textwidth]{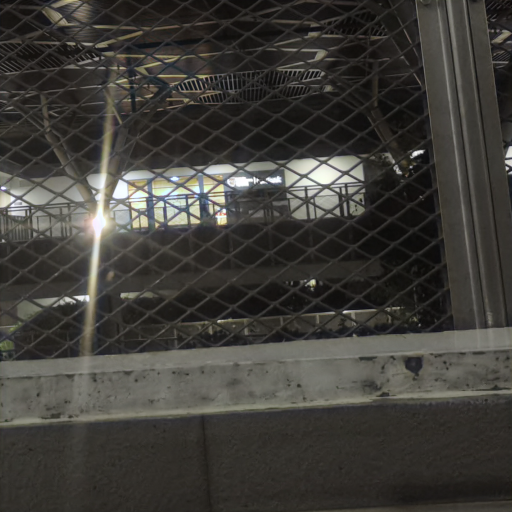}\\
			\includegraphics[width=1\textwidth]{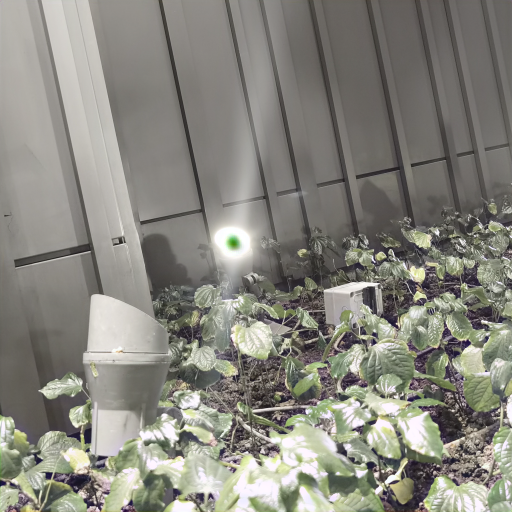}\\ 
			\includegraphics[width=1\textwidth]{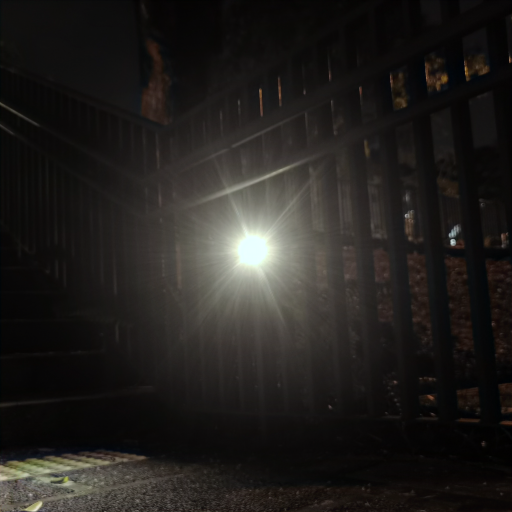}\\ 
			\includegraphics[width=1\textwidth]{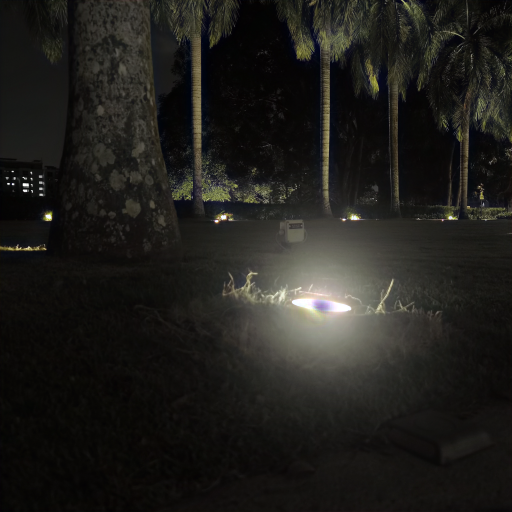}\\
			\includegraphics[width=1\textwidth]{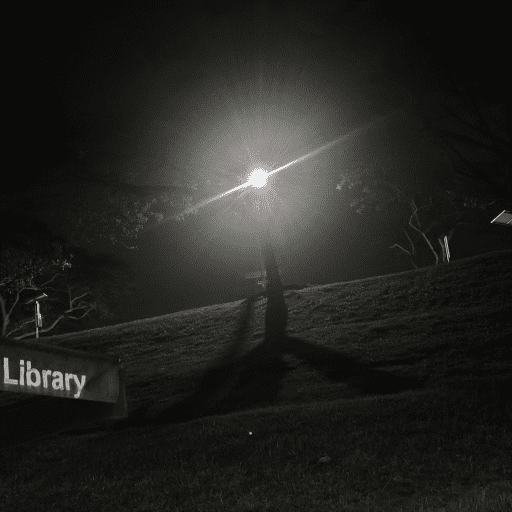}\\ 
			\includegraphics[width=1\textwidth]{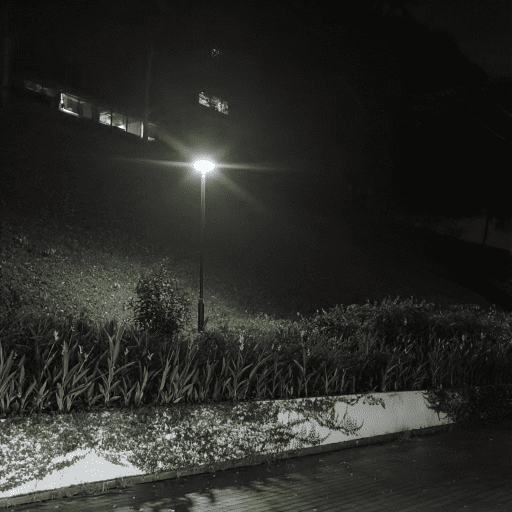}\\ 
			\includegraphics[width=1\textwidth]{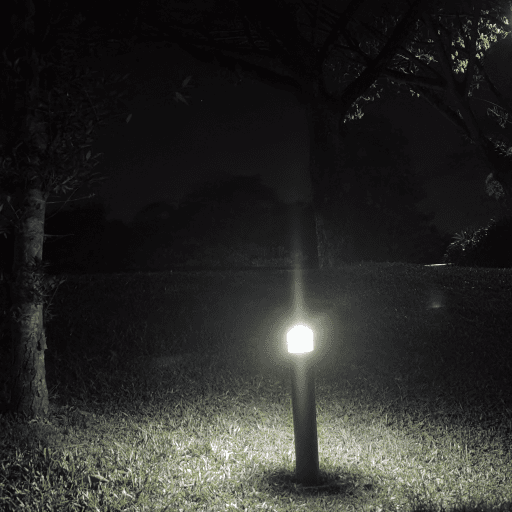}\\ \vspace{-4mm}
		\end{minipage}
	}\hspace{-2mm}
	\subfigure[Wu\cite{wu2021train}]{
		\begin{minipage}[b]{0.134\textwidth}
		    \includegraphics[width=1\textwidth]{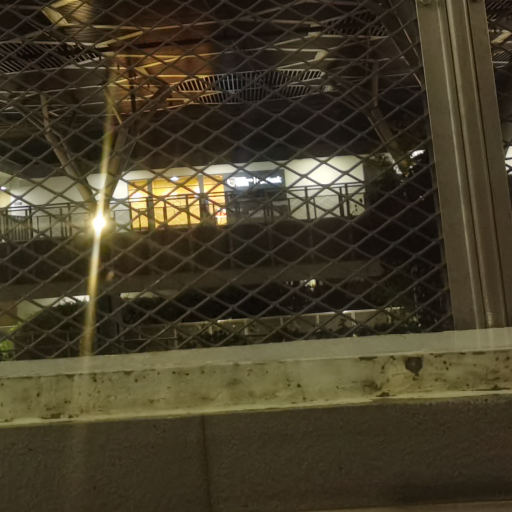}\\
			\includegraphics[width=1\textwidth]{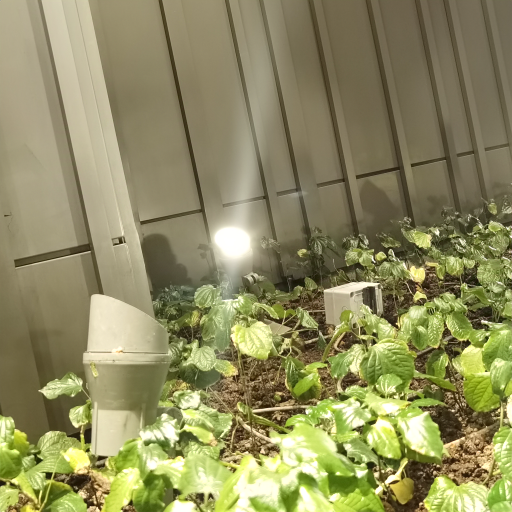}\\ 
			\includegraphics[width=1\textwidth]{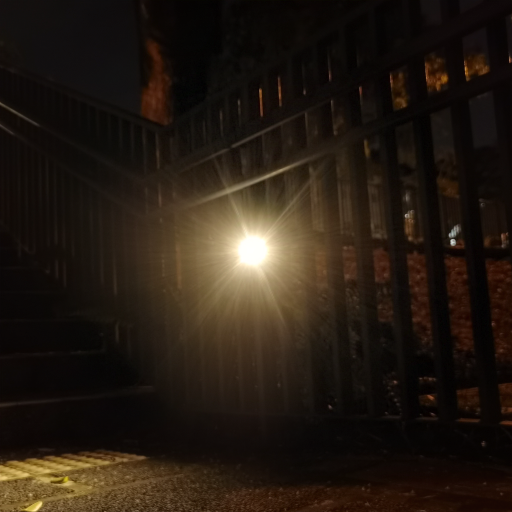}\\ 
			\includegraphics[width=1\textwidth]{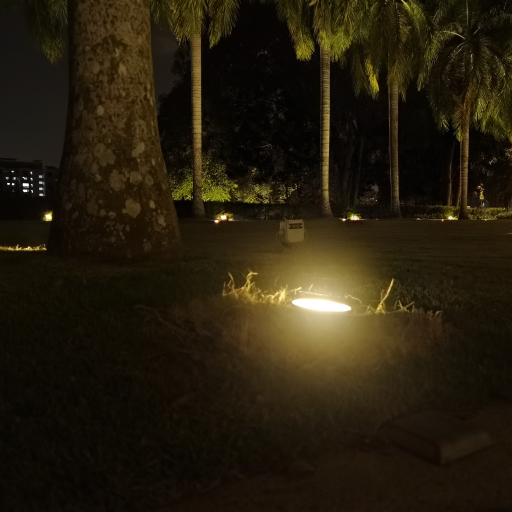}\\
			\includegraphics[width=1\textwidth]{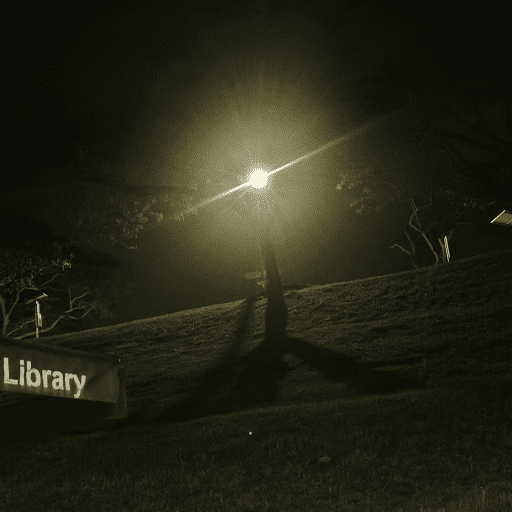}\\ 
			\includegraphics[width=1\textwidth]{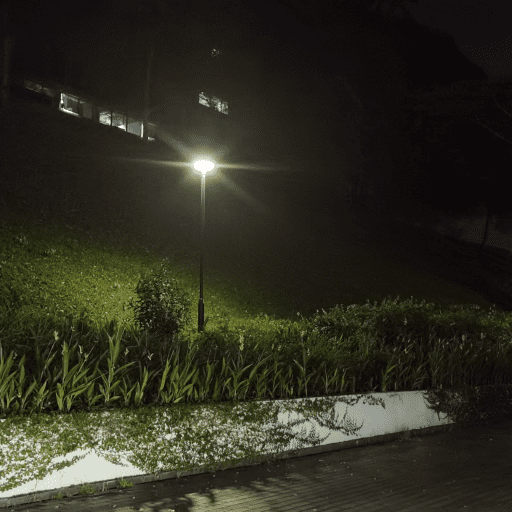}\\ 
			\includegraphics[width=1\textwidth]{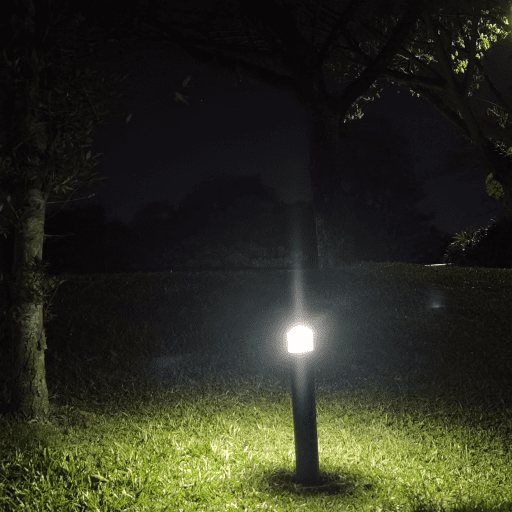}\\ \vspace{-4mm}
		\end{minipage}
	}\hspace{-2mm}
	\subfigure[Dai\cite{dai2022flare7k}]{
		\begin{minipage}[b]{0.134\textwidth}
		    \includegraphics[width=1\textwidth]{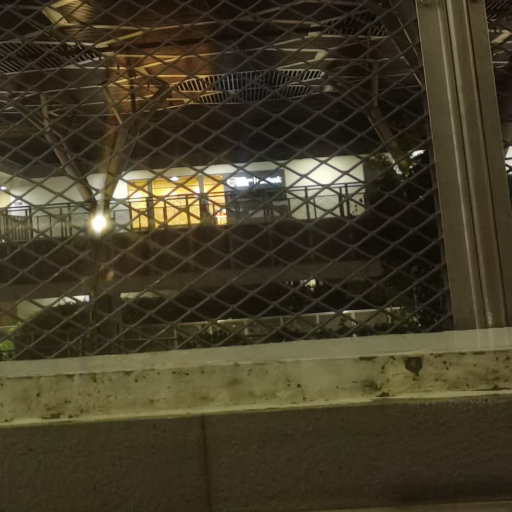}\\
			\includegraphics[width=1\textwidth]{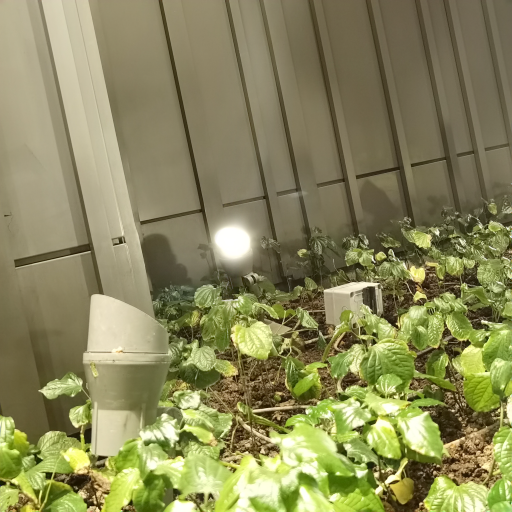}\\ 
			\includegraphics[width=1\textwidth]{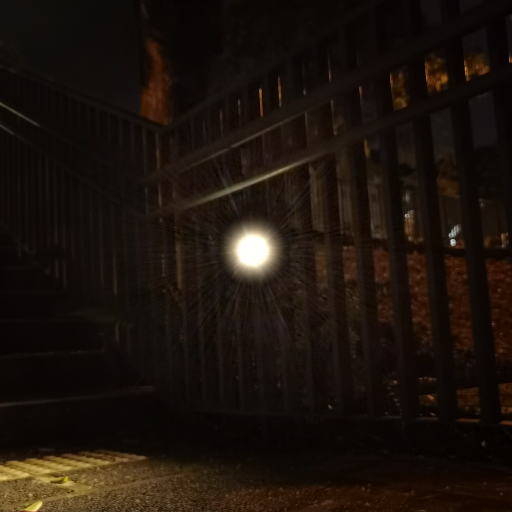}\\ 
			\includegraphics[width=1\textwidth]{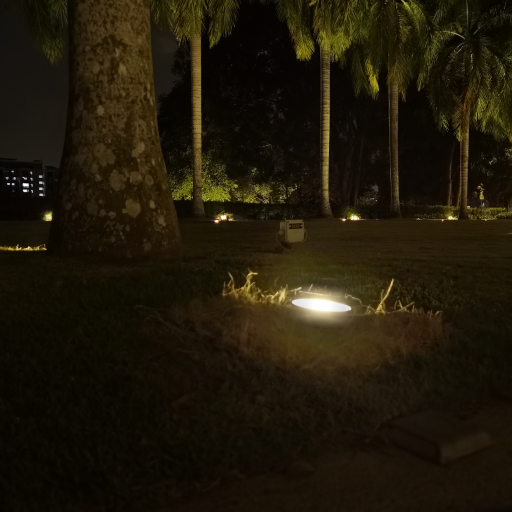}\\
			\includegraphics[width=1\textwidth]{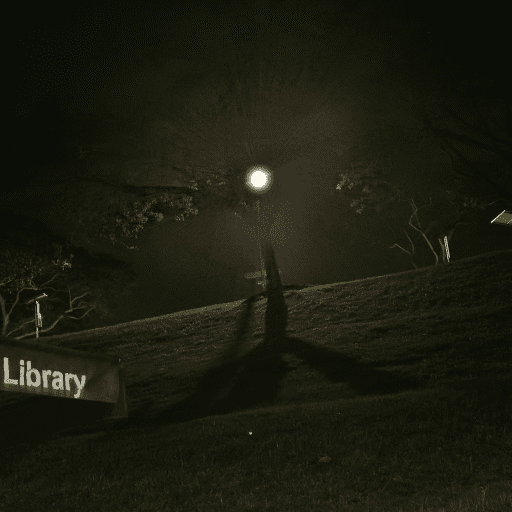}\\ 
			\includegraphics[width=1\textwidth]{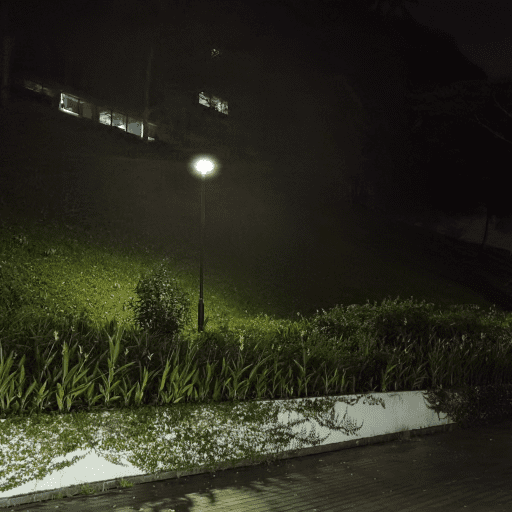}\\ 
			\includegraphics[width=1\textwidth]{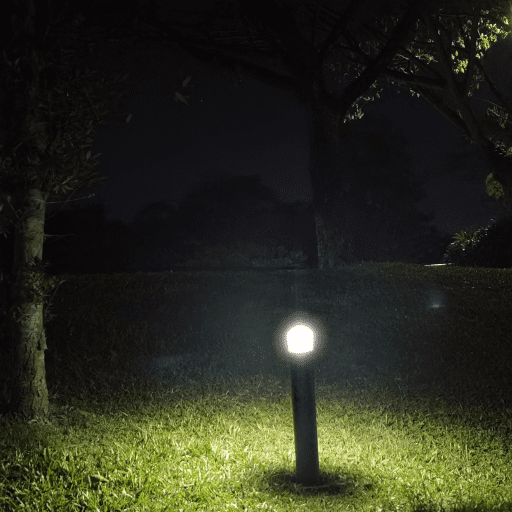}\\ \vspace{-4mm}
		\end{minipage}
	}\hspace{-2mm}
	\subfigure[Uformer\cite{wang2022uformer}]{
		\begin{minipage}[b]{0.134\textwidth}
		    \includegraphics[width=1\textwidth]{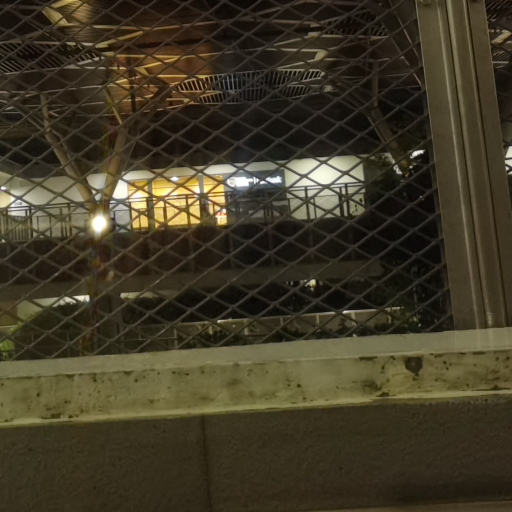}\\
			\includegraphics[width=1\textwidth]{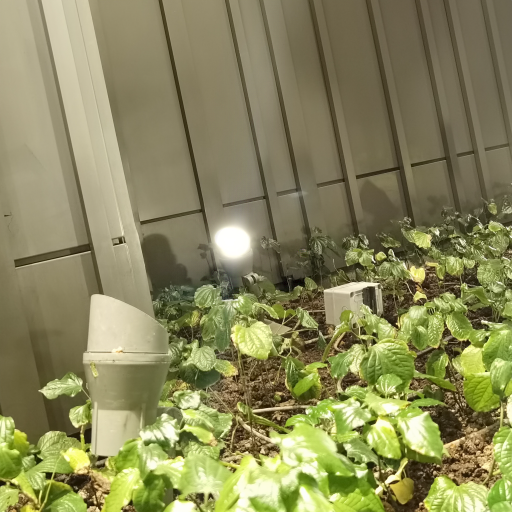}\\ 
			\includegraphics[width=1\textwidth]{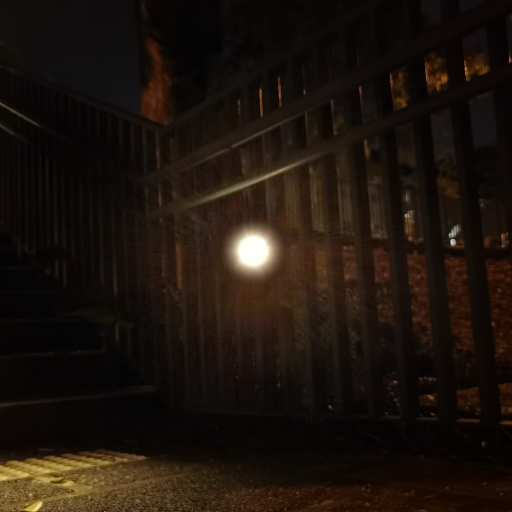}\\ 
			\includegraphics[width=1\textwidth]{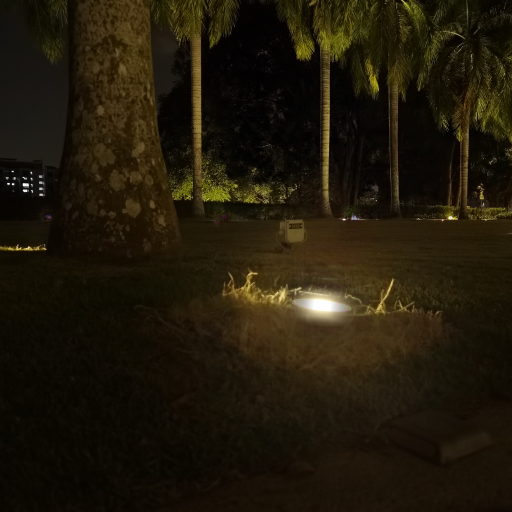}\\
			\includegraphics[width=1\textwidth]{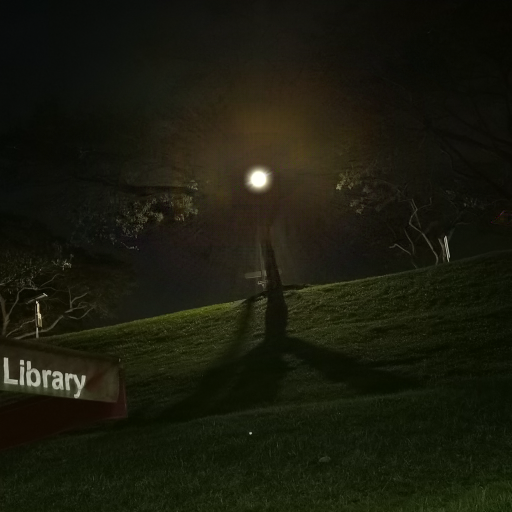}\\ 
			\includegraphics[width=1\textwidth]{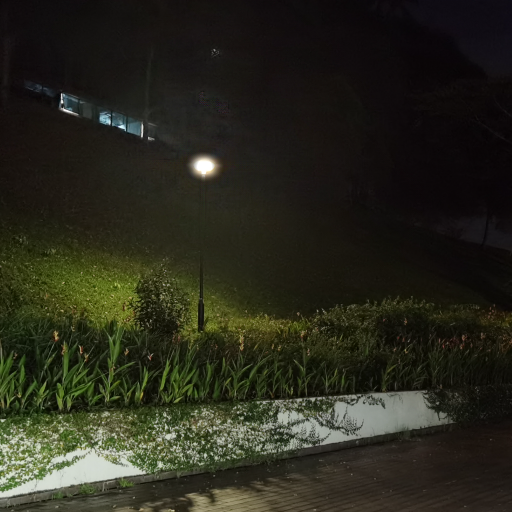}\\ 
			\includegraphics[width=1\textwidth]{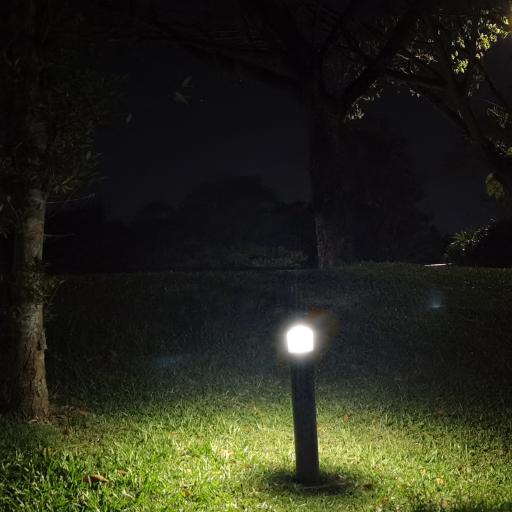}\\ \vspace{-4mm}
		\end{minipage}
	}\hspace{-2mm}
	\subfigure[MFDNet]{
		\begin{minipage}[b]{0.134\textwidth}
		    \includegraphics[width=1\textwidth]{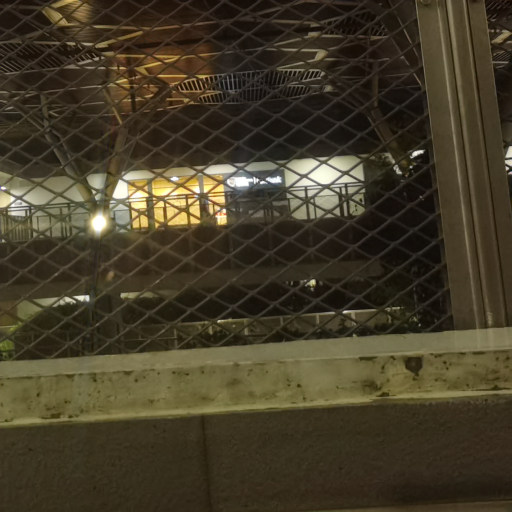}\\
			\includegraphics[width=1\textwidth]{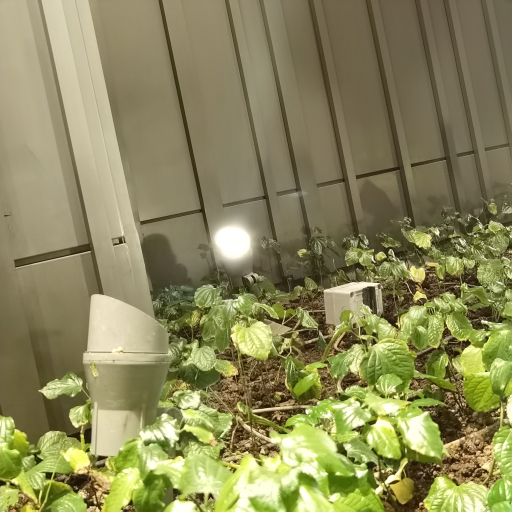}\\ 
			\includegraphics[width=1\textwidth]{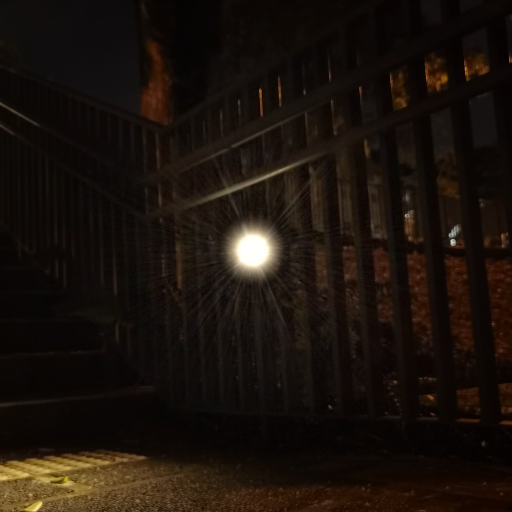}\\ 
			\includegraphics[width=1\textwidth]{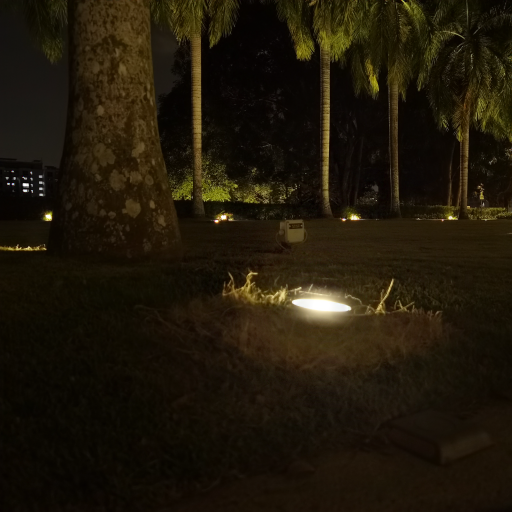}\\
			\includegraphics[width=1\textwidth]{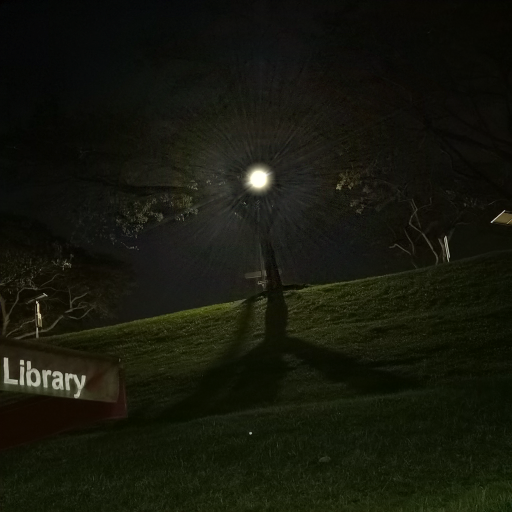}\\ 
			\includegraphics[width=1\textwidth]{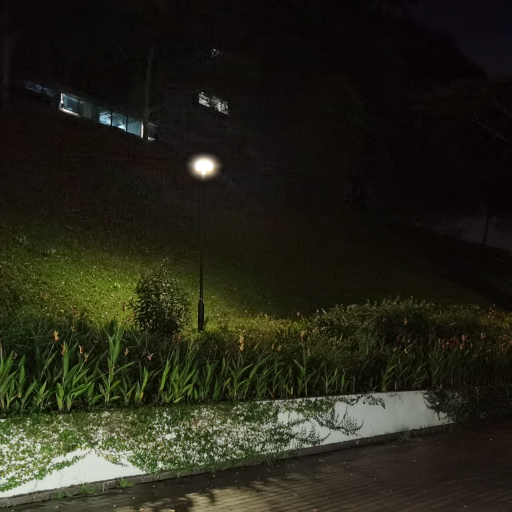}\\ 
			\includegraphics[width=1\textwidth]{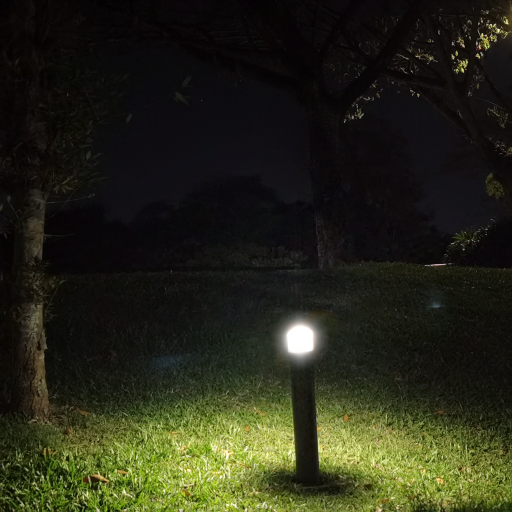}\\ \vspace{-4mm}
		\end{minipage}
	}\hspace{-2mm}
	\subfigure[Ground truth]{
		\begin{minipage}[b]{0.134\textwidth}
   	 	    \includegraphics[width=1\textwidth]{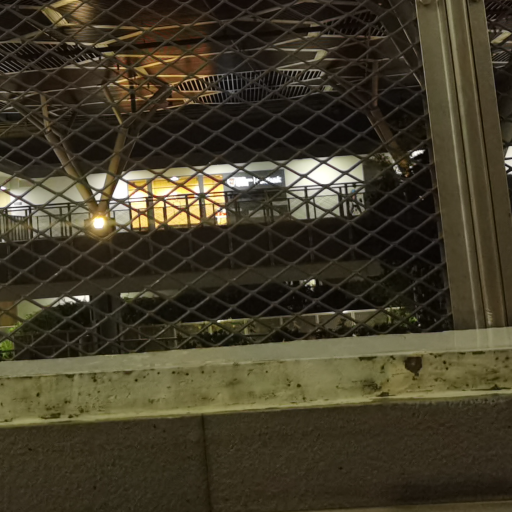}\\
			\includegraphics[width=1\textwidth]{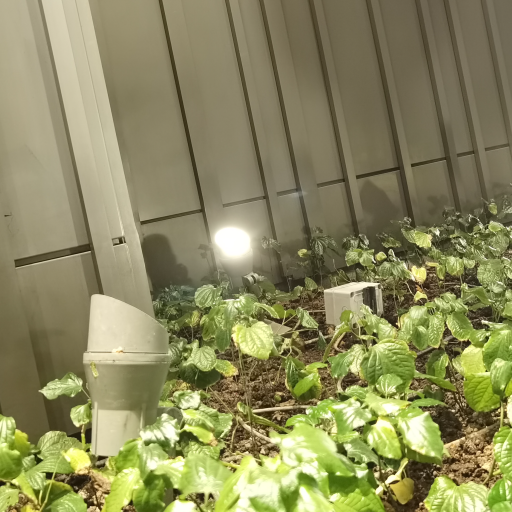}\\ 
			\includegraphics[width=1\textwidth]{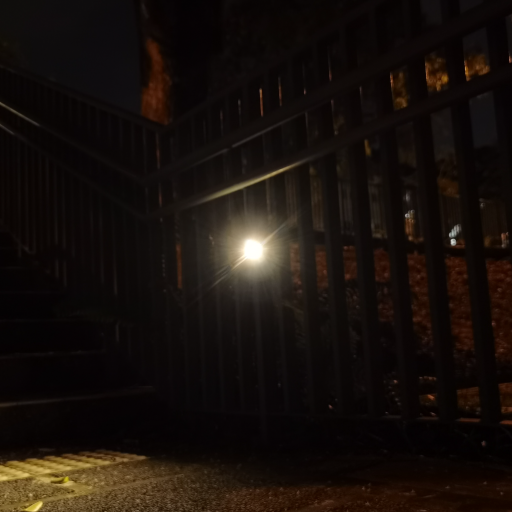}\\ 
			\includegraphics[width=1\textwidth]{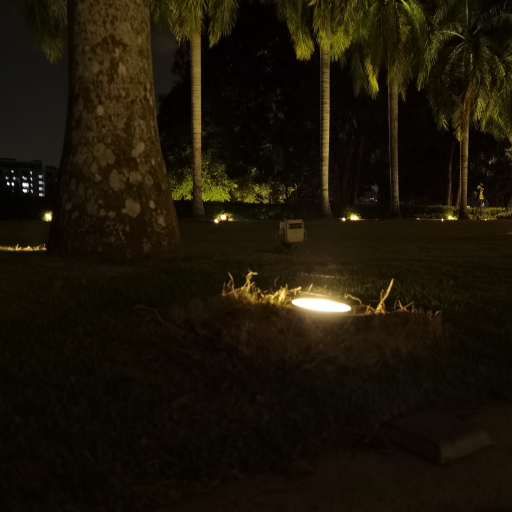}\\
			\includegraphics[width=1\textwidth]{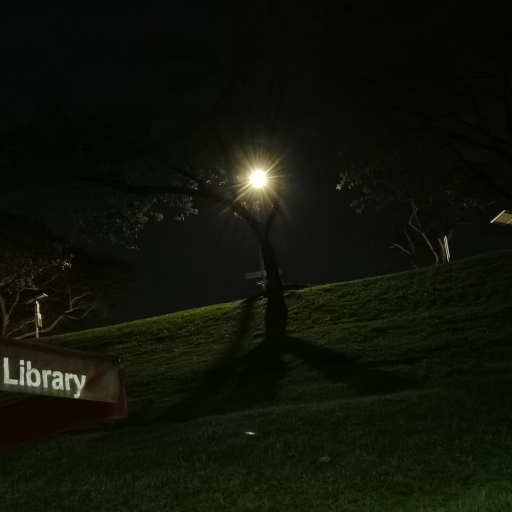}\\ 
			\includegraphics[width=1\textwidth]{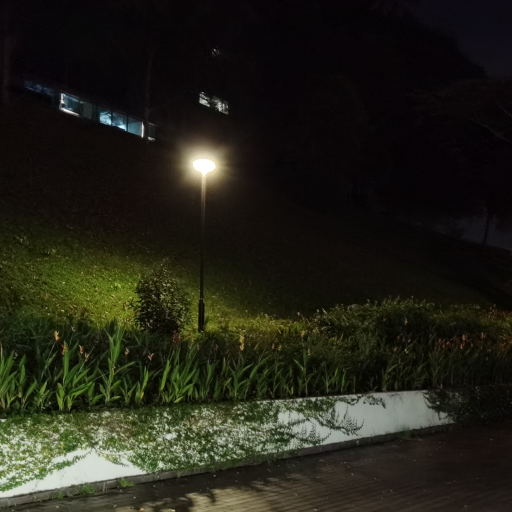}\\ 
			\includegraphics[width=1\textwidth]{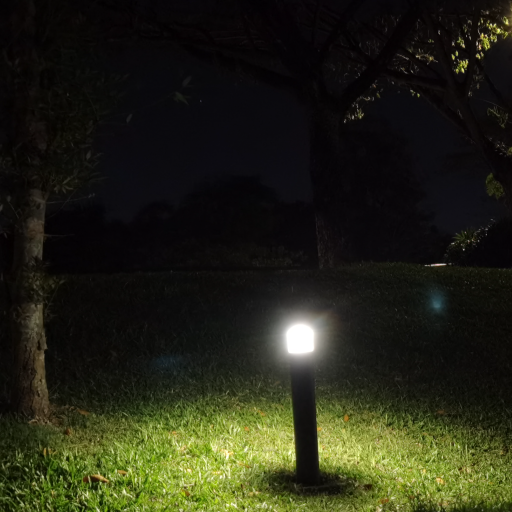}\\ \vspace{-4mm}
		\end{minipage}
		}
	\end{minipage}\vspace{-4mm}
	\caption{Visual comparison on Flare7K real-world nighttime flare-corrupted images.} \vspace{-4mm}
	\label{fig:realmore}
\end{figure*}

\begin{figure*}[!htb]
	\centering
	\begin{minipage}[b]{1.0\textwidth}
	\subfigure[Input]{
		\begin{minipage}[b]{0.134\textwidth}
			\includegraphics[width=1\textwidth]{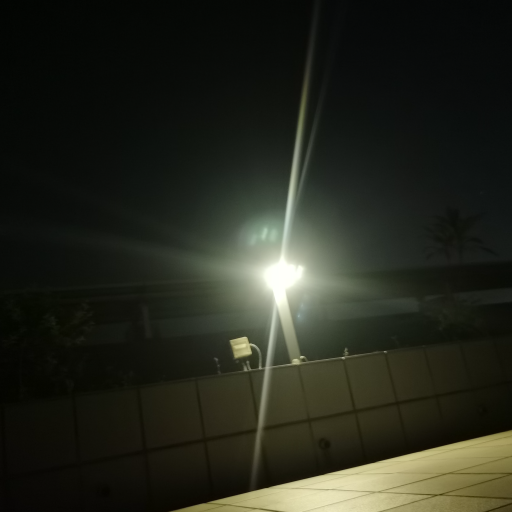}\\ 
			\includegraphics[width=1\textwidth]{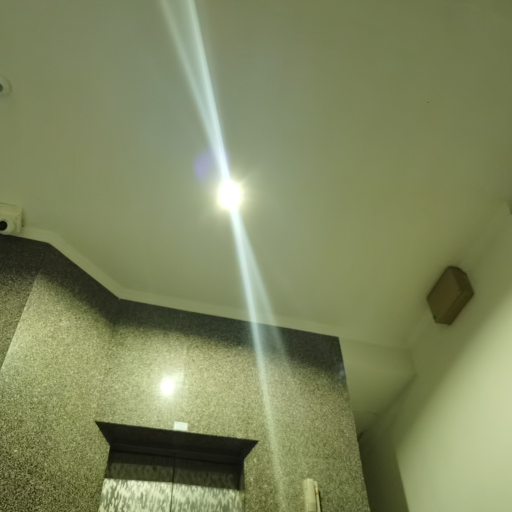}\\ 
			\includegraphics[width=1\textwidth]{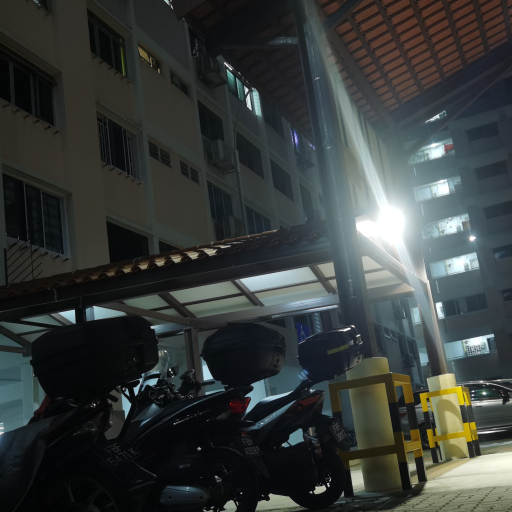}\\ \vspace{-4mm}
		\end{minipage}
	}\hspace{-2mm}
	\subfigure[HINet\cite{chen2021hinet}]{
		\begin{minipage}[b]{0.134\textwidth}
		    \includegraphics[width=1\textwidth]{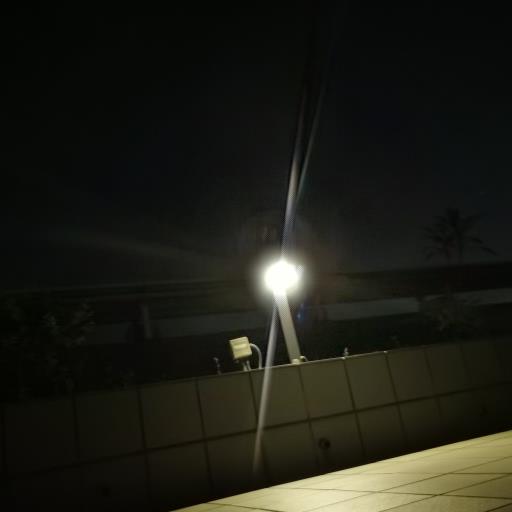}\\ 
		    \includegraphics[width=1\textwidth]{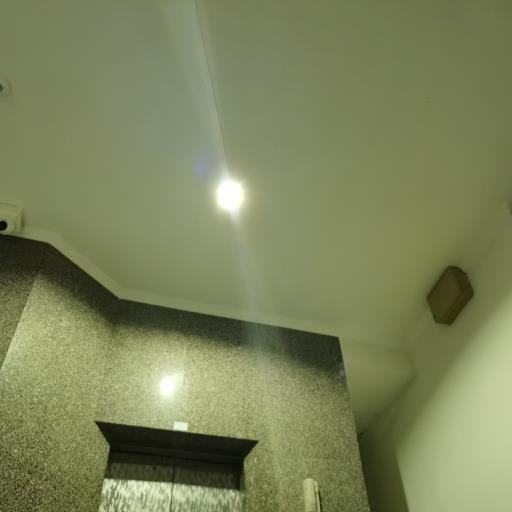}\\ 
		    \includegraphics[width=1\textwidth]{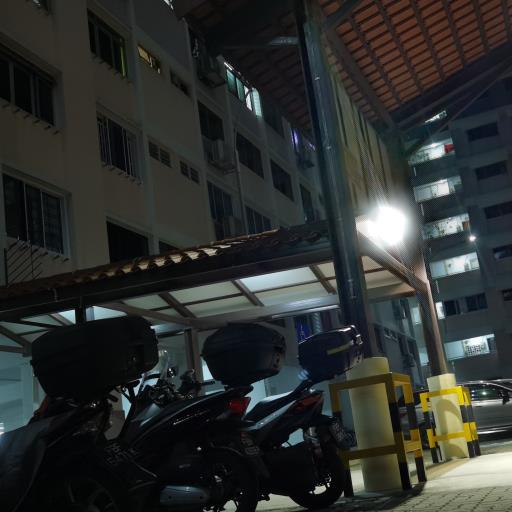}\\ \vspace{-4mm}
		\end{minipage}
	}\hspace{-2mm}
	\subfigure[MPRNet\cite{zamir2021multi}]{
		\begin{minipage}[b]{0.134\textwidth}
		    \includegraphics[width=1\textwidth]{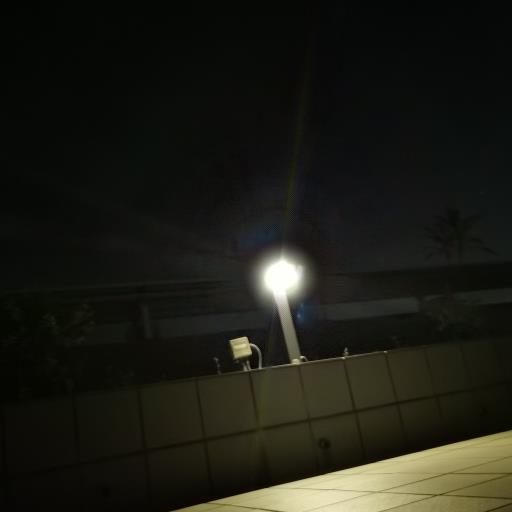}\\
		    \includegraphics[width=1\textwidth]{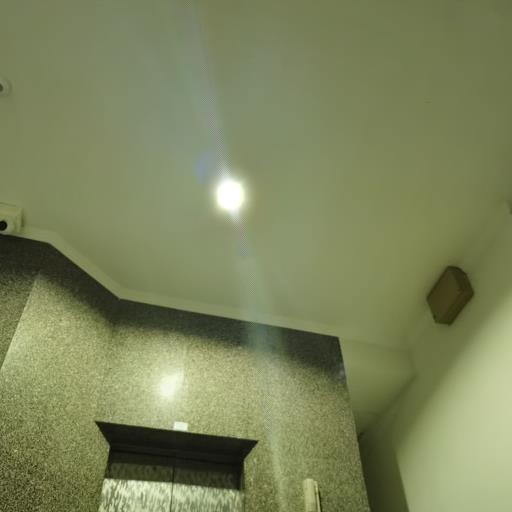}\\ 
		    \includegraphics[width=1\textwidth]{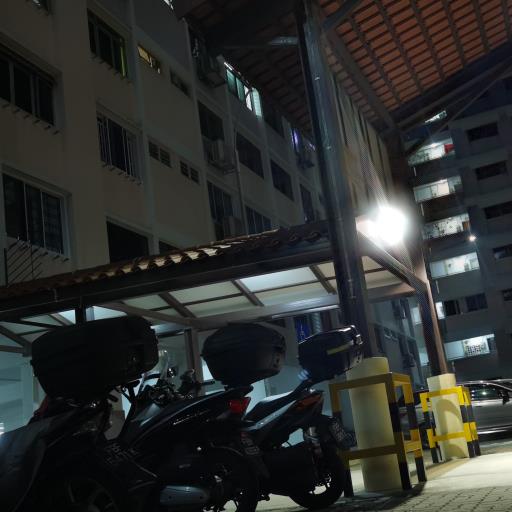}\\ \vspace{-4mm}
		\end{minipage}
	}\hspace{-2mm}
	\subfigure[Restormer\cite{zamir2022restormer}]{
		\begin{minipage}[b]{0.134\textwidth}
		    \includegraphics[width=1\textwidth]{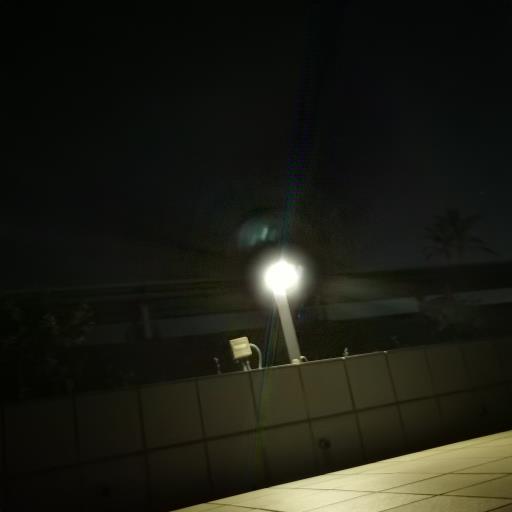}\\ 
		    \includegraphics[width=1\textwidth]{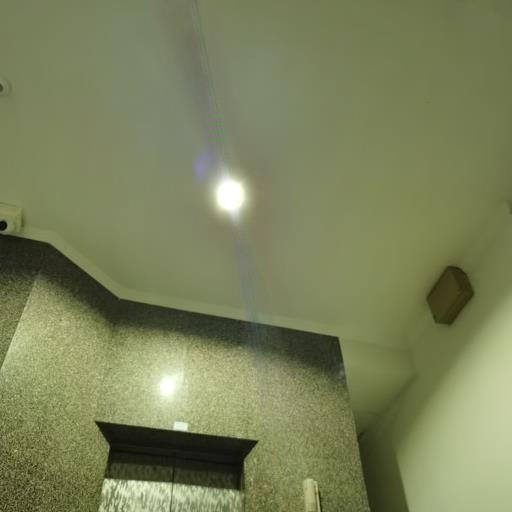}\\
		    \includegraphics[width=1\textwidth]{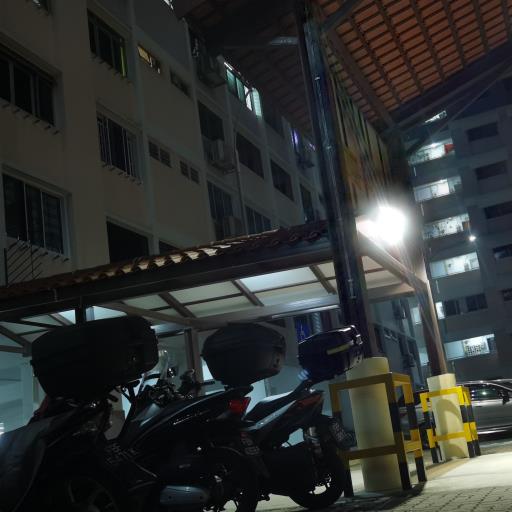}\\\vspace{-4mm}
		\end{minipage}
	}\hspace{-2mm}
	\subfigure[Uformer\cite{wang2022uformer}]{
		\begin{minipage}[b]{0.134\textwidth}
		    \includegraphics[width=1\textwidth]{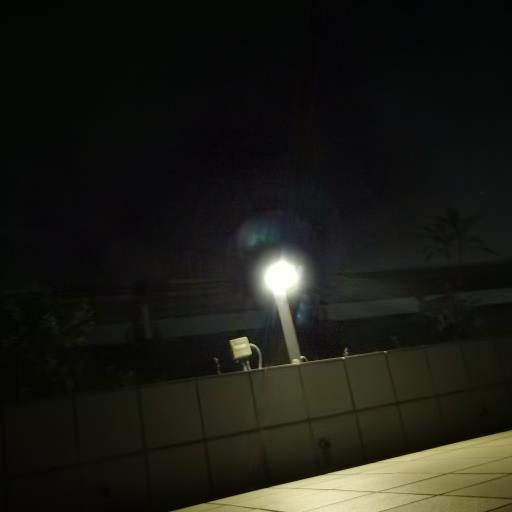}\\ 
		    \includegraphics[width=1\textwidth]{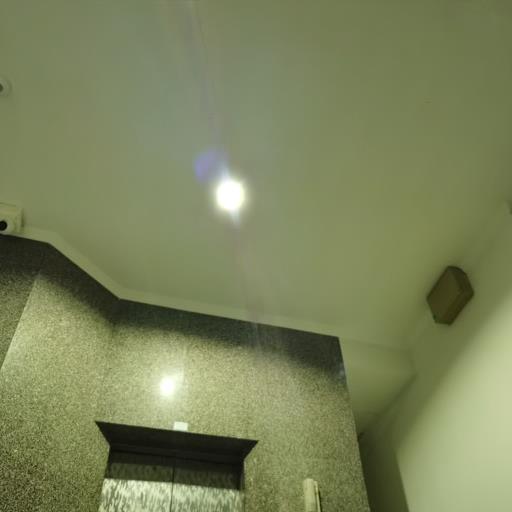}\\ 
		    \includegraphics[width=1\textwidth]{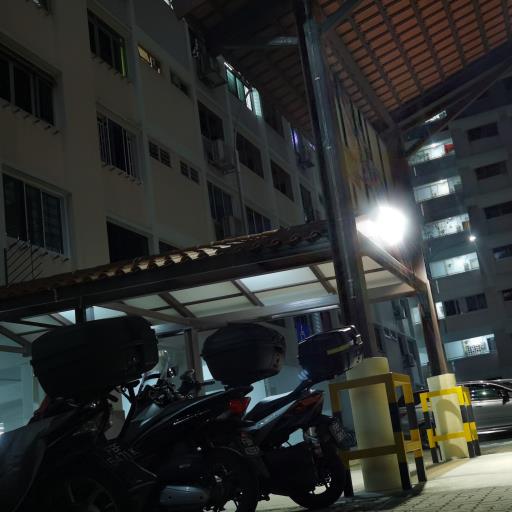}\\ \vspace{-4mm}
		\end{minipage}
	}\hspace{-2mm}
	\subfigure[MFDNet]{
		\begin{minipage}[b]{0.134\textwidth}
		    \includegraphics[width=1\textwidth]{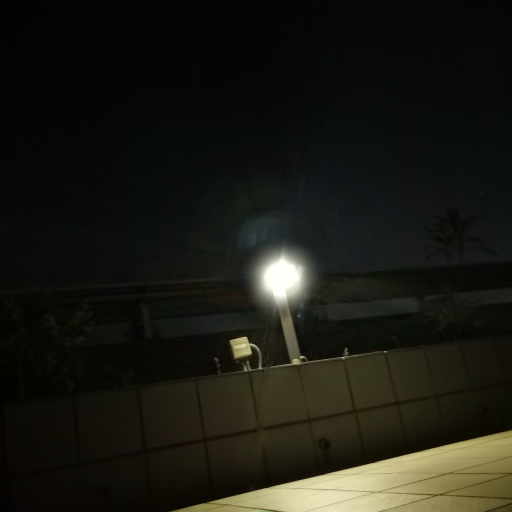}\\ 
		    \includegraphics[width=1\textwidth]{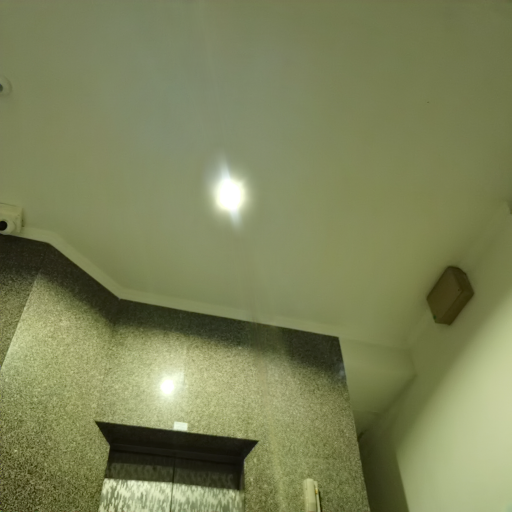}\\ 
		    \includegraphics[width=1\textwidth]{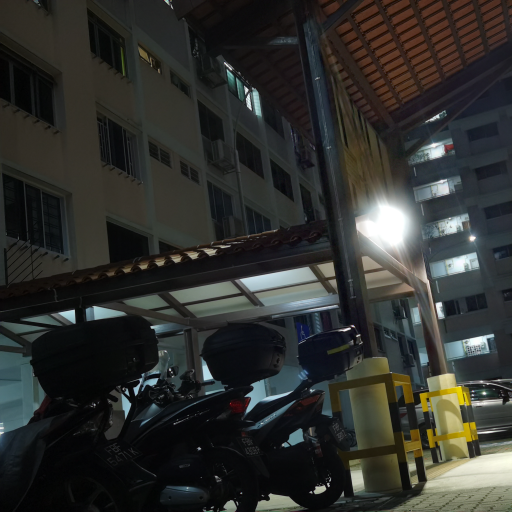}\\ \vspace{-4mm}
		\end{minipage}
	}\hspace{-2mm}
	\subfigure[Ground truth]{
		\begin{minipage}[b]{0.134\textwidth}
   	 	    \includegraphics[width=1\textwidth]{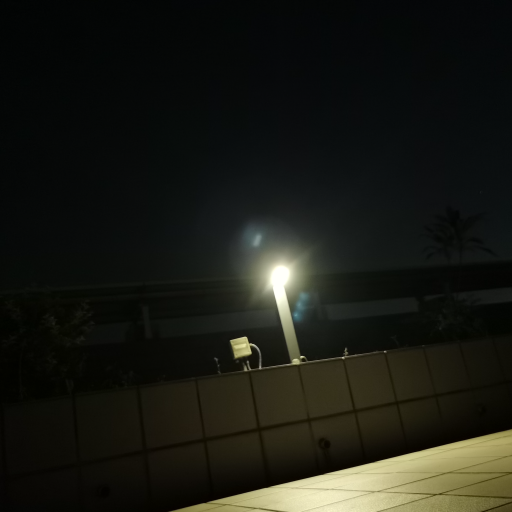}\\ 
   	 	    \includegraphics[width=1\textwidth]{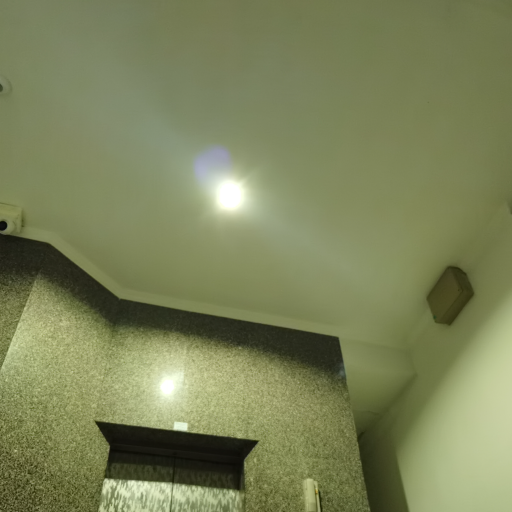}\\ 
   	 	    \includegraphics[width=1\textwidth]{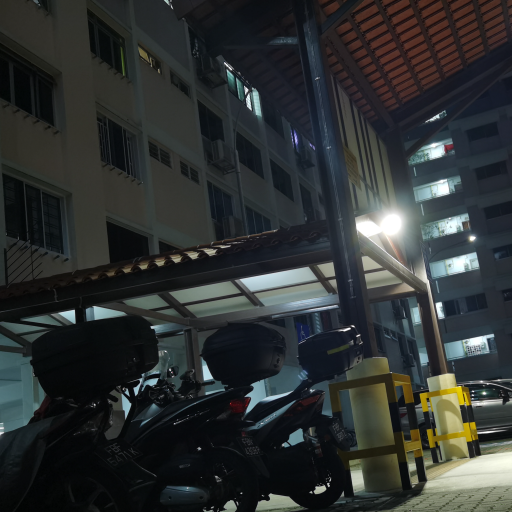}\\ \vspace{-4mm}
		\end{minipage}
		}
	\end{minipage}\vspace{-4mm}
	\caption{Visual comparison on Flare7K real-world nighttime flare-corrupted images.} \vspace{-4mm}
	\label{fig:real}
\end{figure*}

\section{Experiments and Results} \label{experiment}

\subsection{Setup}

\subsubsection{Dataset}

We train our MFDNet based on the Flare7K dataset \cite{dai2022flare7k}. Flare7K is currently the largest publicly available nighttime flare dataset, consisting of 5,000 scattering and 2,000 reflective flare images. The dataset comprises 25 types of scattering flares and 10 types of reflective flares. We combine the Flare7K flare image with 23949 flare-free images sampled from the 24K Flick images \cite{zhang2018single} to create paired flare-corrupted and flare-free images for training. Furthermore, Flare7K provides 100 real-world flare-corrupted images and 100 synthetic flare-corrupted images, along with their corresponding ground truth, which serve as the test dataset.


\subsubsection{Implementation Details}

Our implementation is based on PyTorch. We use the Adam optimizer with a learning rate of $1e-4$.  For a fair comparison, we use the same data augmentation strategy and post-processing step as \cite{dai2022flare7k}. To augment the training samples, we use random rotation, translation, shear, scale, and flip transformations. During the post-processing stage, we extract the saturated regions of the input image and superimpose it back onto the deflared image to recover the light source. In addition, our MFDNet is designed as a scalable model and the number of LP's decomposed layers $n$ is related to the size of the input image. According to \cite{liang2021high,ijcai2023p129,10378157}, increasing $n$ reduces the computational burden but degrades the performance. As the number of LP's decomposed layers $n$ increases, the image size of the lowest frequency part decreases, which can reduce the computational cost. However, as $n$ increases, the amount of effective information in the lowest-frequency components diminishes, which impacts the low-frequency deflaring and the quality of the reconstructed image. Consequently, the model's performance deteriorates with increasing depth. We set up $n=3$ to get a trade-off between the computational cost and the performance.

\subsubsection{Evaluation Metrics}
We utilize various metrics to evaluate our MFDNet's performance. Specifically, we measure the quality of nighttime flare removal using Structural Similarity (SSIM) \cite{wang2004image}, Peak Signal-to-Noise Ratio (PSNR), and Learned Perceptual Image Patch Similarity (LPIPS) \cite{zhang2018unreasonable}. Additionally, we measure the method's practicality using GMACs, parameters, and inference time.

\subsection{Comparison with the State-of-the-art Methods}
We compare the performance of MFDNet with several state-of-the-art methods for nighttime flare removal in Flare7K real-world and synthetic datasets. These methods include a nighttime dehazing method \cite{zhang2020nighttime}, a nighttime visual enhancement method \cite{sharma2021nighttime}, two U-Net-based flare removal methods \cite{wu2021train,dai2022flare7k}, and HINet \cite{chen2021hinet}, MPRNet \cite{zamir2021multi}, Restormer \cite{zamir2022restormer}, Uformer \cite{wang2022uformer}.

\subsubsection{Qualitative Evaluation}
\begin{figure*}[!htb]
	\centering
	\begin{minipage}[b]{1.0\textwidth}
	\subfigure[Input]{
		\begin{minipage}[b]{0.134\textwidth}
			\includegraphics[width=1\textwidth]{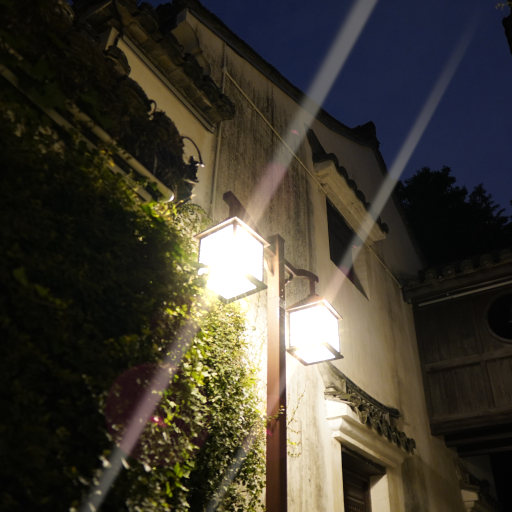}\\
			\includegraphics[width=1\textwidth]{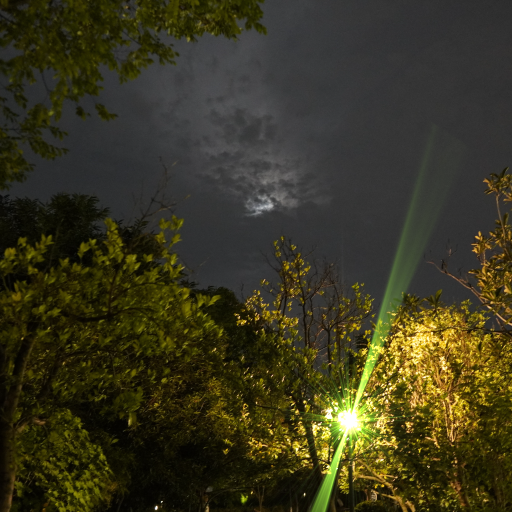}\\ 
			\includegraphics[width=1\textwidth]{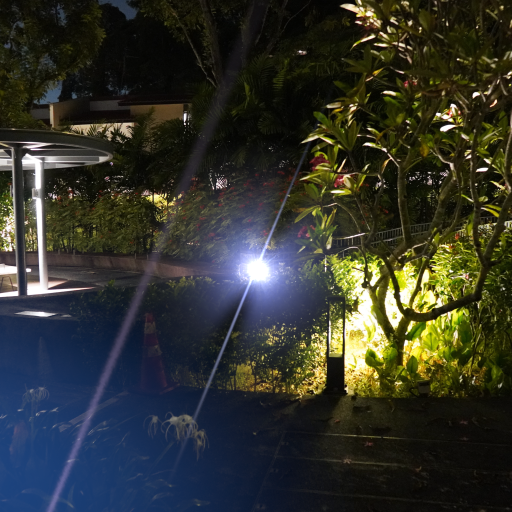}\\ 
			\includegraphics[width=1\textwidth]{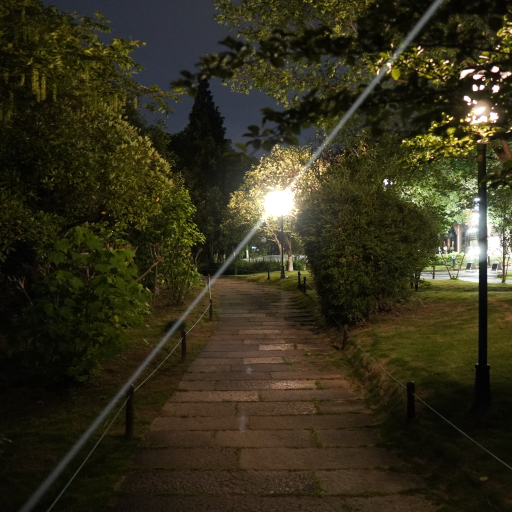}\\ 
			\includegraphics[width=1\textwidth]{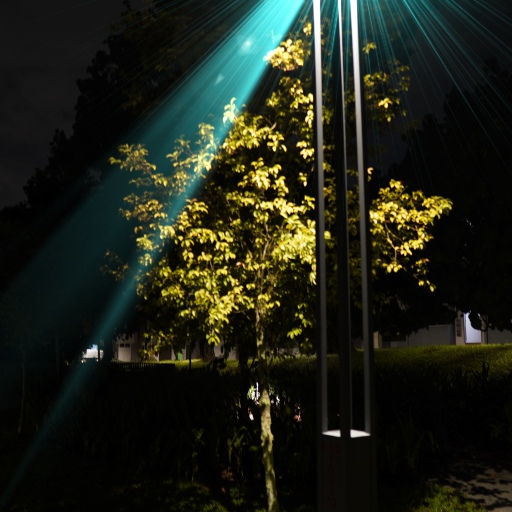}\\ 
			\includegraphics[width=1\textwidth]{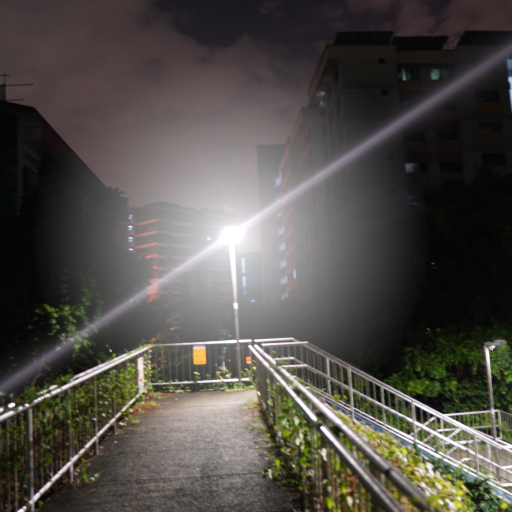}\\ \vspace{-4mm}
		\end{minipage}
	}\hspace{-2mm}
	\subfigure[Zhang\cite{zhang2020nighttime}]{
		\begin{minipage}[b]{0.134\textwidth}
		    \includegraphics[width=1\textwidth]{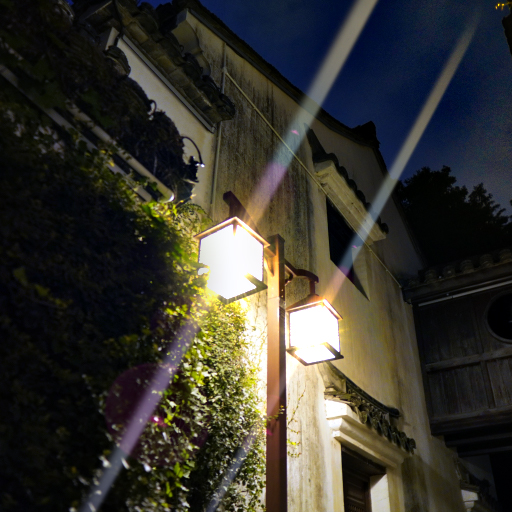}\\ 
		    \includegraphics[width=1\textwidth]{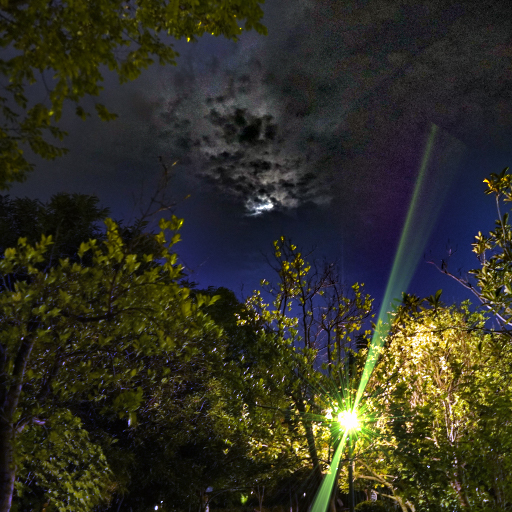}\\ 
		    \includegraphics[width=1\textwidth]{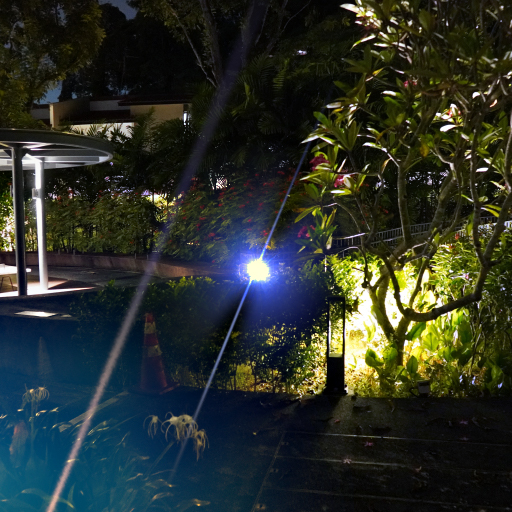}\\ 
		    \includegraphics[width=1\textwidth]{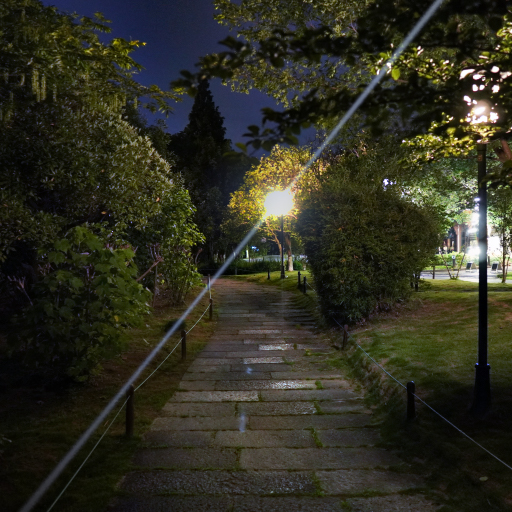}\\ 
		    \includegraphics[width=1\textwidth]{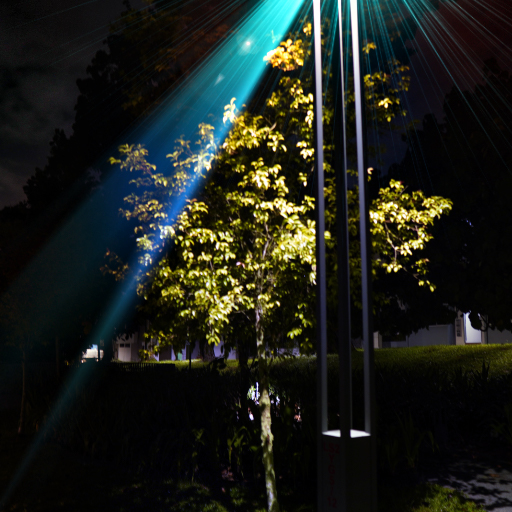}\\ 
		    \includegraphics[width=1\textwidth]{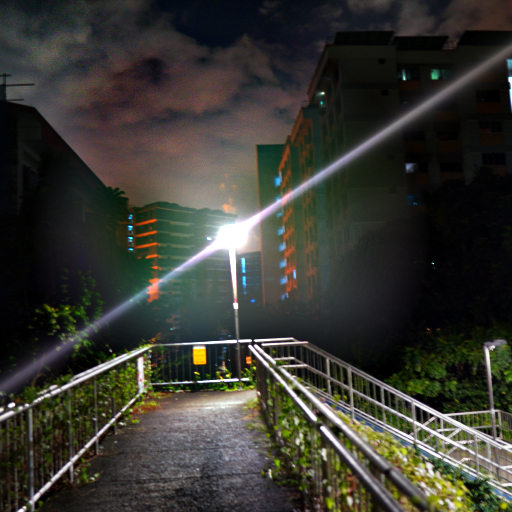}\\ \vspace{-4mm}
		\end{minipage}
	}\hspace{-2mm}
	\subfigure[Sharma\cite{sharma2021nighttime}]{
		\begin{minipage}[b]{0.134\textwidth}
		    \includegraphics[width=1\textwidth]{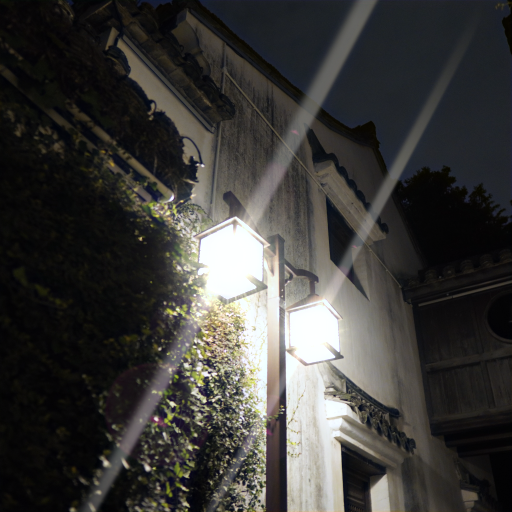}\\
		    \includegraphics[width=1\textwidth]{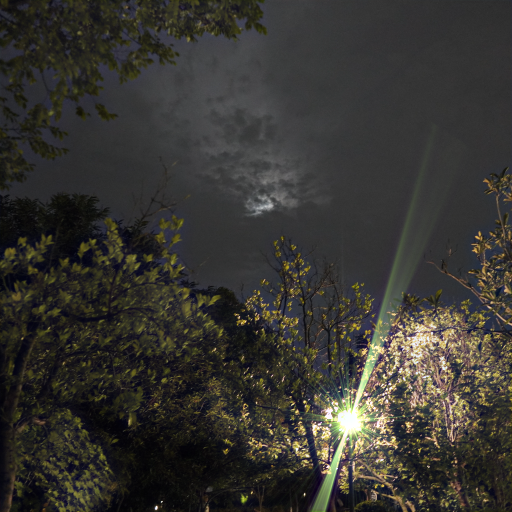}\\ 
		    \includegraphics[width=1\textwidth]{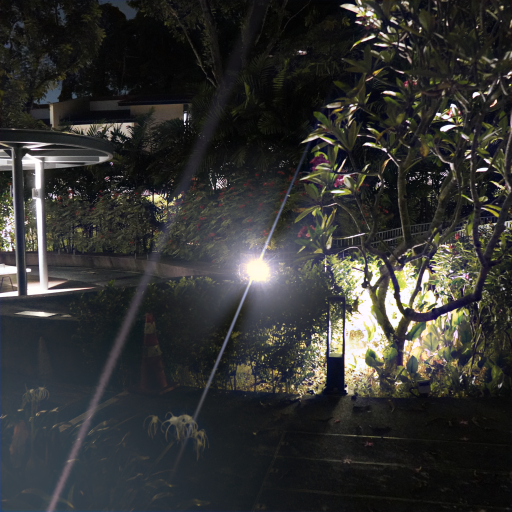}\\ 
		    \includegraphics[width=1\textwidth]{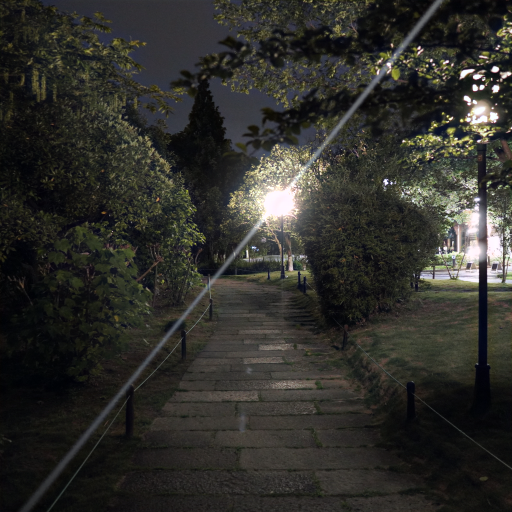}\\ 
		    \includegraphics[width=1\textwidth]{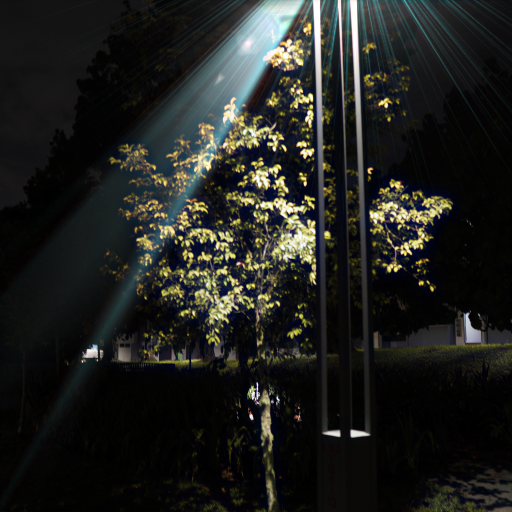}\\ 
		    \includegraphics[width=1\textwidth]{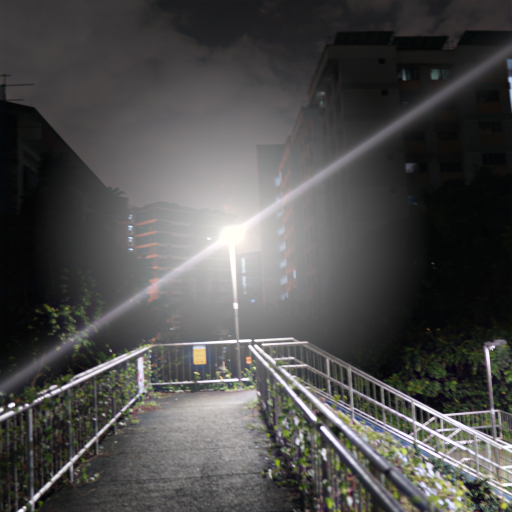}\\ \vspace{-4mm}
		\end{minipage}
	}\hspace{-2mm}
	\subfigure[Wu\cite{wu2021train}]{
		\begin{minipage}[b]{0.134\textwidth}
		    \includegraphics[width=1\textwidth]{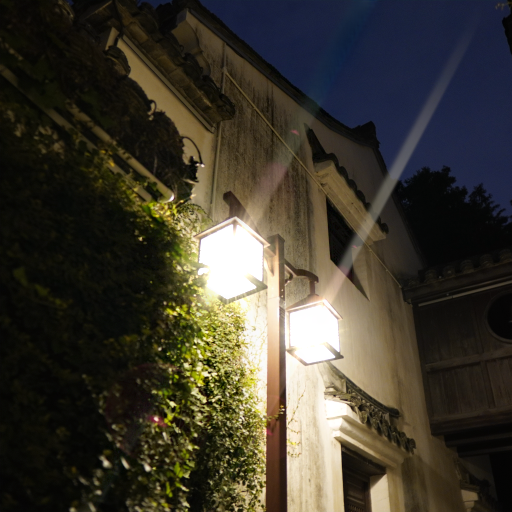}\\ 
		    \includegraphics[width=1\textwidth]{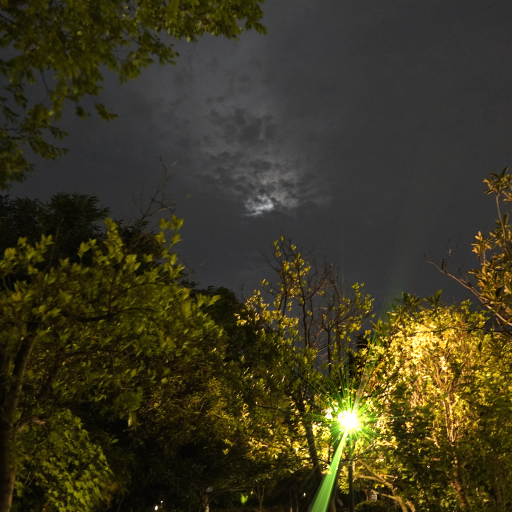}\\
		    \includegraphics[width=1\textwidth]{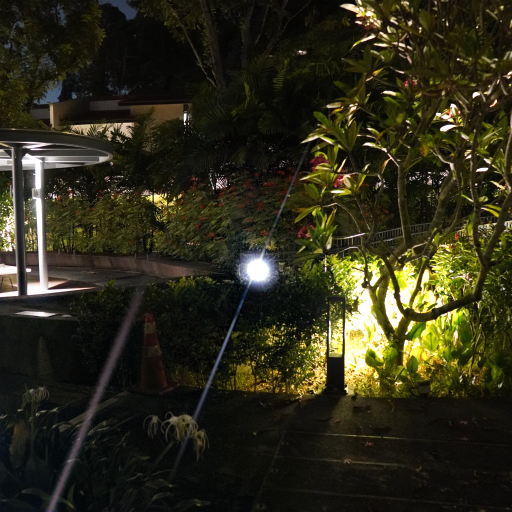}\\
		    \includegraphics[width=1\textwidth]{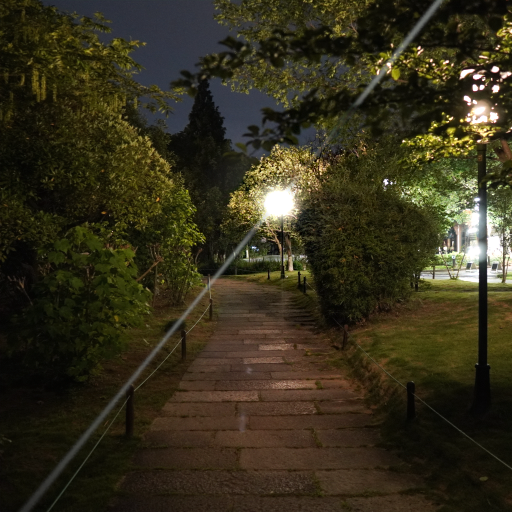}\\
		    \includegraphics[width=1\textwidth]{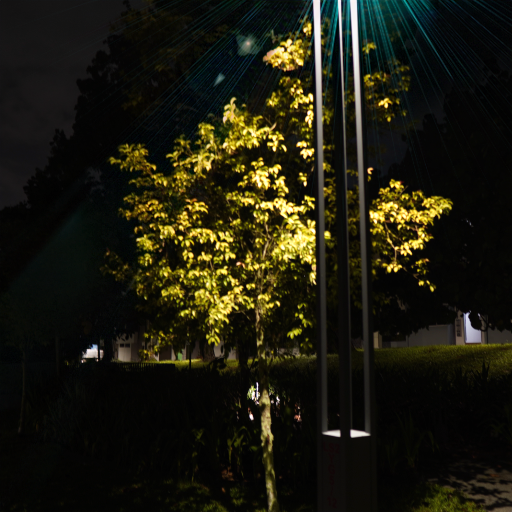}\\
		    \includegraphics[width=1\textwidth]{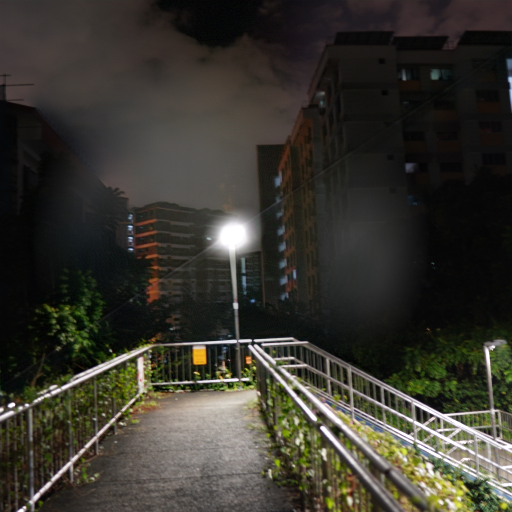}\\ \vspace{-4mm}
		\end{minipage}
	}\hspace{-2mm}
	\subfigure[Dai\cite{dai2022flare7k}]{
		\begin{minipage}[b]{0.134\textwidth}
		    \includegraphics[width=1\textwidth]{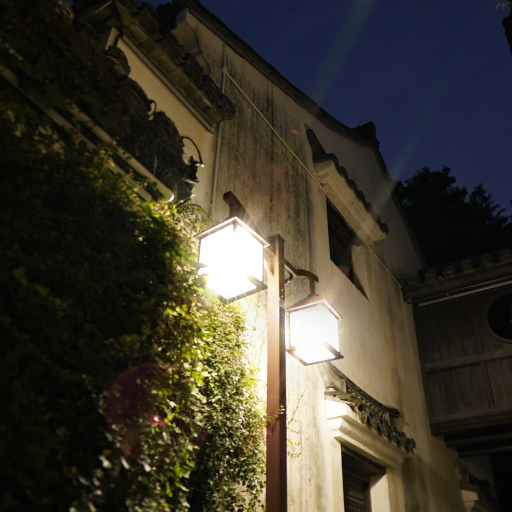}\\ 
		    \includegraphics[width=1\textwidth]{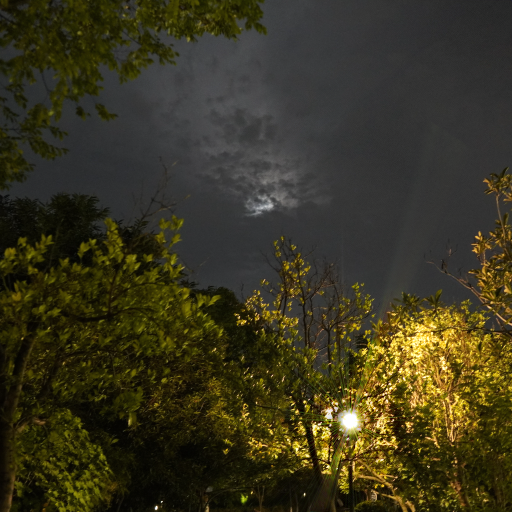}\\ 
		    \includegraphics[width=1\textwidth]{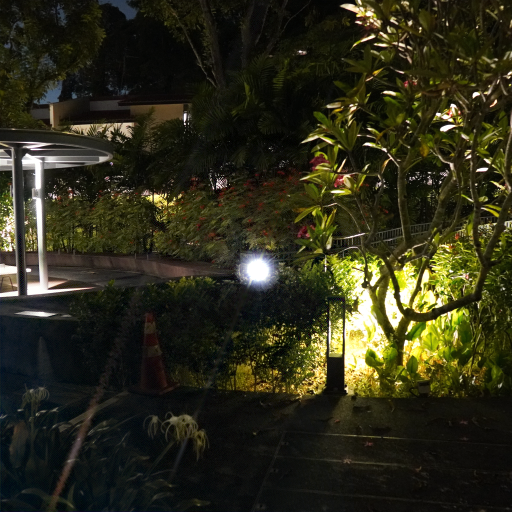}\\ 
		    \includegraphics[width=1\textwidth]{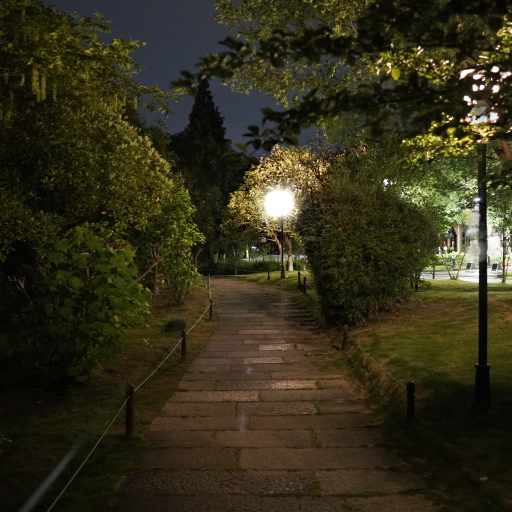}\\ 
		    \includegraphics[width=1\textwidth]{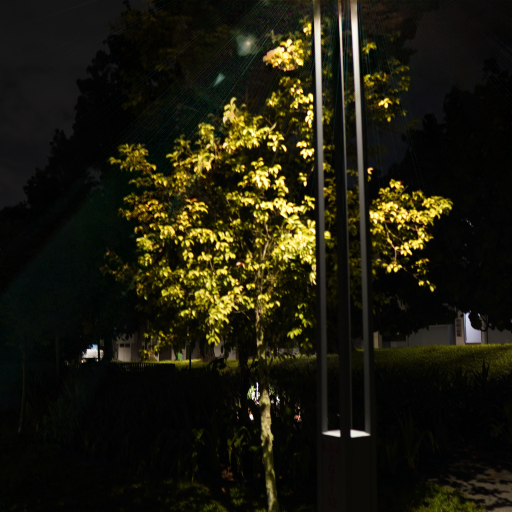}\\ 
		    \includegraphics[width=1\textwidth]{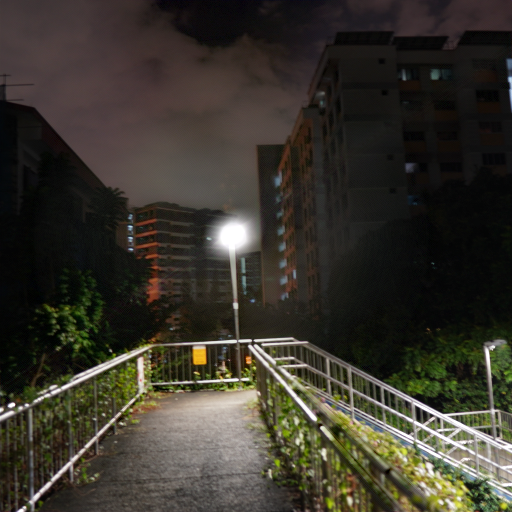}\\ \vspace{-4mm}
		\end{minipage}
	}\hspace{-2mm}
	\subfigure[MFDNet]{
		\begin{minipage}[b]{0.134\textwidth}
		    \includegraphics[width=1\textwidth]{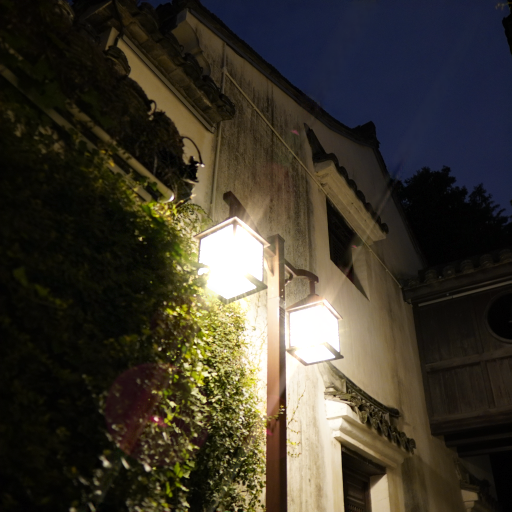}\\ 
		    \includegraphics[width=1\textwidth]{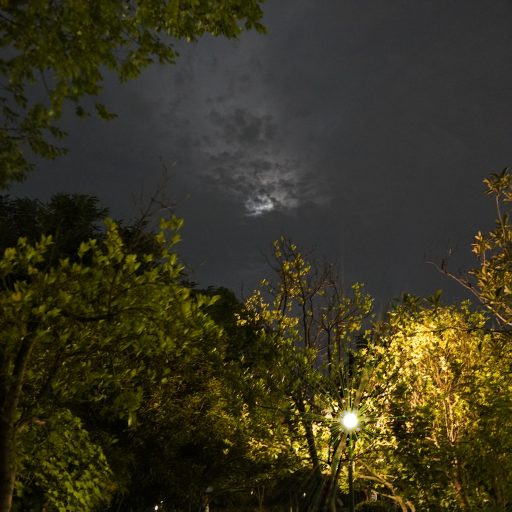}\\ 
		    \includegraphics[width=1\textwidth]{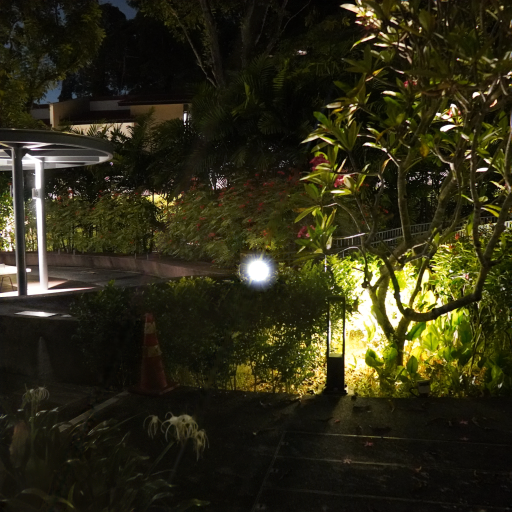}\\ 
		    \includegraphics[width=1\textwidth]{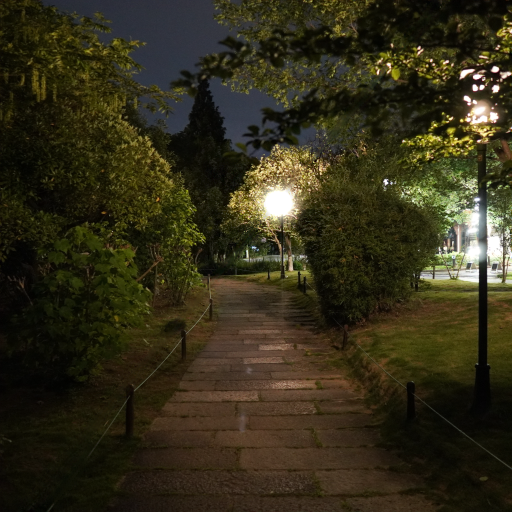}\\ 
		    \includegraphics[width=1\textwidth]{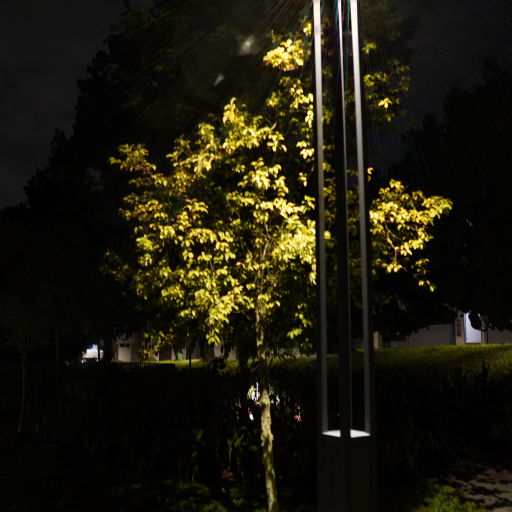}\\ 
		    \includegraphics[width=1\textwidth]{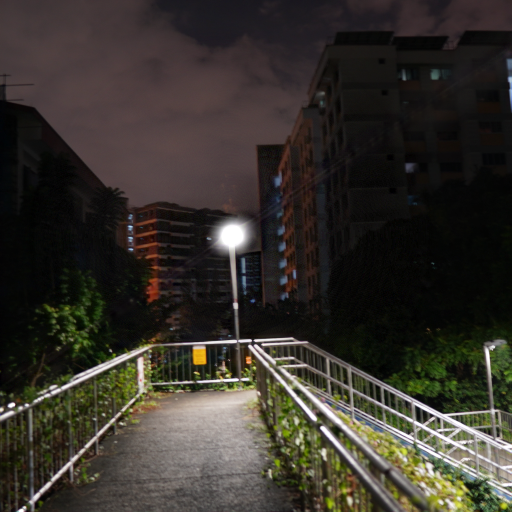}\\ \vspace{-4mm}
		\end{minipage}
	}\hspace{-2mm}
	\subfigure[Ground truth]{
		\begin{minipage}[b]{0.134\textwidth}
   	 	    \includegraphics[width=1\textwidth]{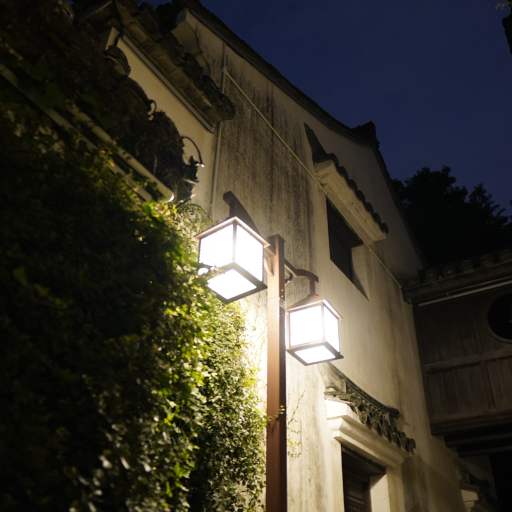}\\
   	 	    \includegraphics[width=1\textwidth]{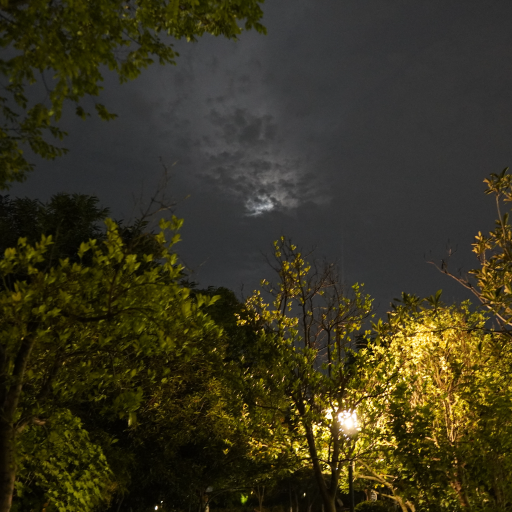}\\ 
   	 	    \includegraphics[width=1\textwidth]{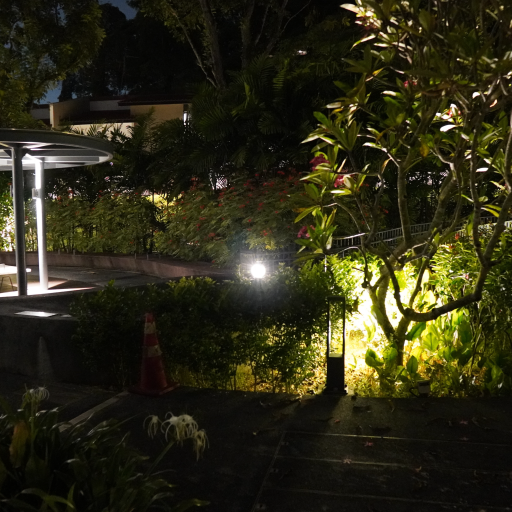}\\ 
   	 	    \includegraphics[width=1\textwidth]{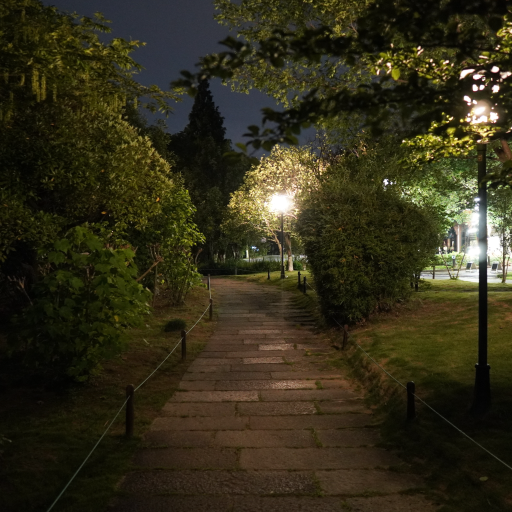}\\ 
   	 	    \includegraphics[width=1\textwidth]{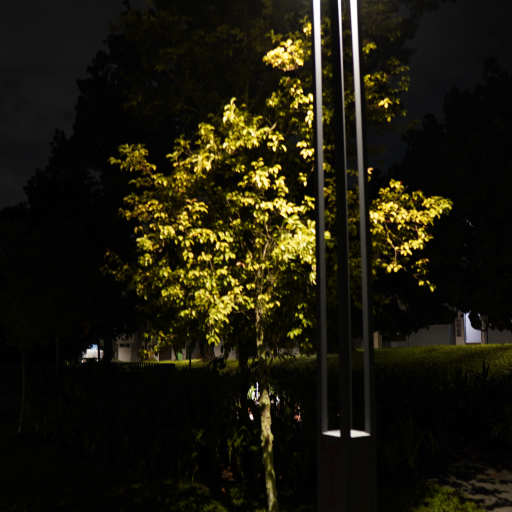}\\ 
   	 	    \includegraphics[width=1\textwidth]{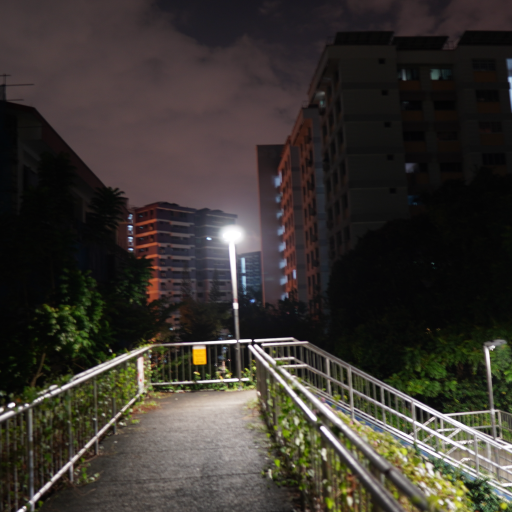}\\  \vspace{-4mm}
		\end{minipage}
		}\hspace{-2mm}
	\end{minipage}\hspace{-2mm}\vspace{-4mm}
	\caption{Visual comparison on Flare7K synthetic nighttime flare-corrupted images.} 
	\label{fig:more_comparison_synthetic}
\end{figure*}

We first compare the visual results of real-world nighttime flare removal in Figure \ref{fig:realmore} and Figure \ref{fig:real}. The visualization results show that the flare-free images recovered by our MFDNet are closer to the ground-truth images. Subsequently, we provide a visual comparison of removing synthetic flares in Figure \ref{fig:more_comparison_synthetic}. Our MFDNet still restores cleaner flare-free images than those of the other algorithms. As shown in Figure \ref{fig:realmore}, Figure \ref{fig:real}, and Figure \ref{fig:more_comparison_synthetic}, the previous methods could not completely remove the flare, or alter the original color and texture details of the input image while removing the flare. Our method can remove the flare while preserving the original image information.

Specifically, to further prove the effectiveness and robustness of our method, we compare the flare removal results of our MFDNet and other state-of-the-art methods for different flare patterns. In the fourth row of Figure \ref{fig:realmore}, the input image contains the glare artifact, in the sixth row of Figure \ref{fig:realmore}, the input image contains the shimmer flare, in the second row of Figure \ref{fig:real}, the input image contains the streak flare, in the fifth row of Figure \ref{fig:more_comparison_synthetic}, the input image contains colored lines. It can be seen that for these different flare patterns, the deflared image recovered through our method is closer to the ground truth. At the same time, our MFDNet can remove multiple flare patterns that exist within a single image, such as the third row of Figure \ref{fig:realmore}, which contains both shimmer and glare artifacts, the first row of Figure \ref{fig:real}, and the last row of Figure \ref{fig:more_comparison_synthetic}, which contain both streak and glare artifacts.

\begin{table*}[!ht]
\begin{center}
\caption{Comparison of the computational cost and time consumption on images from $512\times512$ resolution to 4K resolution. 'N.A.' denotes that the method cannot handle the input image of this size, and 'OOM' means that the method causes the out-of-memory issue for this specific resolution. The best results are highlighted in bold.} 
\adjustbox{width=\linewidth}{
\begin{tabular}{c | c c c  | c c c  | c c c  | c c c  | c c c  }
\toprule[0.15em]
   & \multicolumn{3}{c|}{\textbf{512$\times$512}} & \multicolumn{3}{c|}{\textbf{1024$\times$1024}} & \multicolumn{3}{c|}{\textbf{1080p}} & \multicolumn{3}{c|}{\textbf{2K}} & \multicolumn{3}{c}{\textbf{4K}} \\
\cline{2-16}
   Method  &GMACs & Params& Time~&GMACs& Params& Time ~ &GMACs & Params& Time~ &GMACs & Params& Time~&GMACs & Params& Time~  \\
   ~&(G)~ & (M)~& (s)~ &(G)~ & (M)~& (s)~ &(G)~ & (M)~& (s)~ &(G)~ & (M)~& (s)~ &(G)~ & (M)~& (s)~   \\
\midrule[0.15em]
Dai~\cite{dai2022flare7k}  & 261.90 & 34.53 & 0.052 & 1047.61 & 34.53 & 0.195 & N.A.&N.A.&N.A.&3682.99 & 34.53 & 0.686 & {OOM}&{OOM}&{OOM} \\
HINet~\cite{chen2021hinet} & 682.86 & 88.67 & 0.119 & 2731.43 & 88.67 & 0.448 & N.A.&N.A.&N.A.& {OOM}&{OOM}&{OOM} & {OOM}&{OOM}&{OOM} \\
MPRNet~\cite{zamir2021multi}& 274.97 & \textbf{1.93} & 0.101 & 1099.87 & \textbf{1.93} & 0.377 & 2175.04&\textbf{1.93}&0.760&{OOM}&{OOM}&{OOM} & {OOM}&{OOM}&{OOM}  \\
Restormer~\cite{zamir2022restormer}& 57.61 & 2.98 & 0.106 & 230.46 & 2.98 & 0.394 & 455.74&2.98&0.771&{OOM}&{OOM}&{OOM} & {OOM}&{OOM}&{OOM} \\
Uformer~\cite{wang2022uformer} & 160.09 & 20.43 & 0.137 & 640.35 & 20.43 & 0.536 &N.A.&N.A.&N.A.&N.A.&N.A.&N.A.&N.A.&N.A.&N.A.\\

\textbf{MFDNet (ours)} &\textbf{18.25}  & 6.32&\textbf{0.034} & \textbf{73.01} & 6.32&\textbf{0.075}  & \textbf{144.53}  & 6.32&\textbf{0.153} & \textbf{256.66}  & \textbf{6.32}&\textbf{0.301}  & \textbf{578.12}  & \textbf{6.32}  & \textbf{0.783}   \\
\bottomrule[0.15em]
\end{tabular}
}\vspace{-4mm}
\label{table:efficiency}
\end{center}
\end{table*}

\begin{figure*}[!htb]
	\centering
	\begin{minipage}[b]{1.0\textwidth}
	\subfigure[Input]{
		\begin{minipage}[b]{0.49\textwidth}
			\includegraphics[width=1\textwidth]{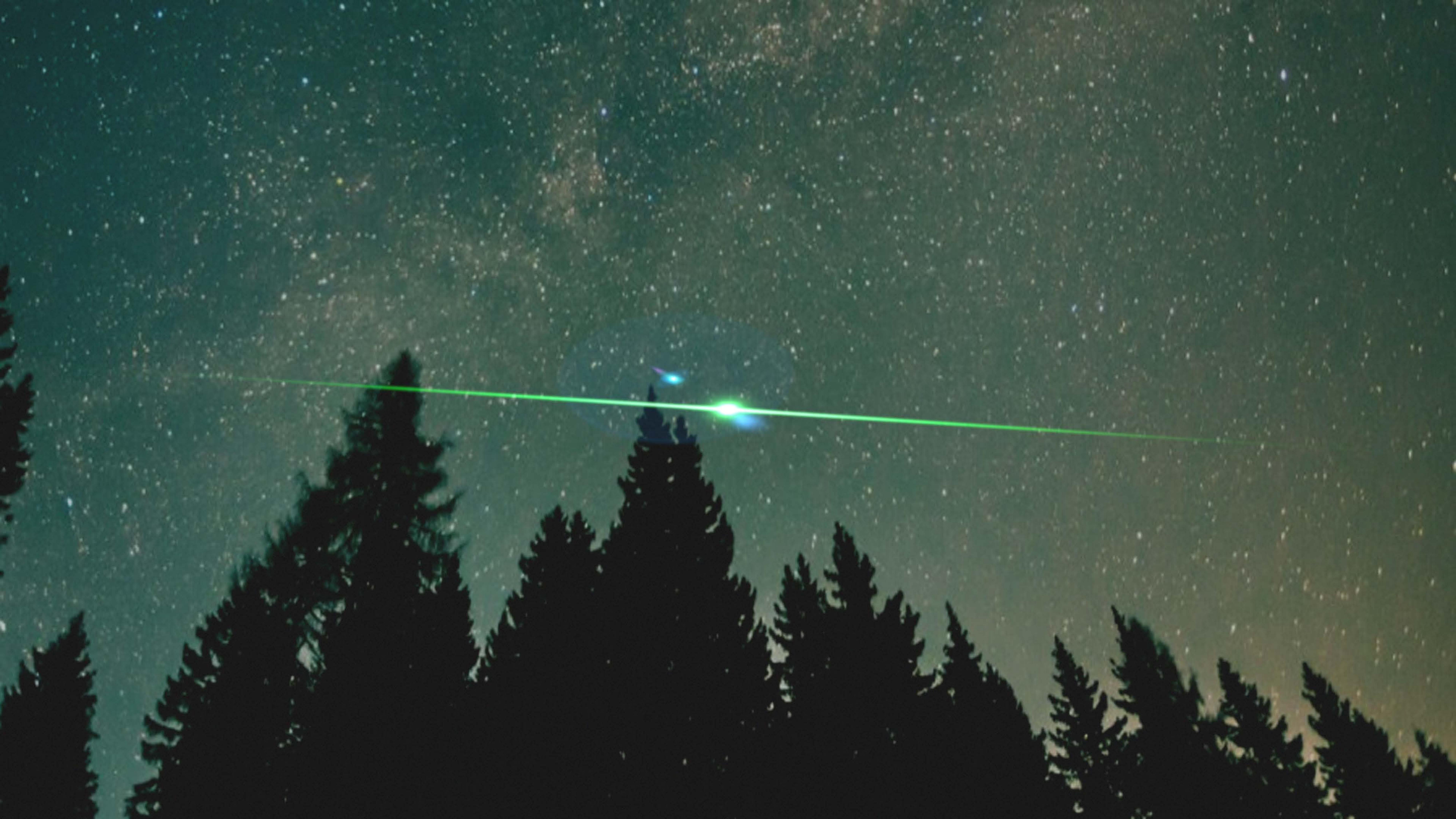}\\ 
			\includegraphics[width=1\textwidth]{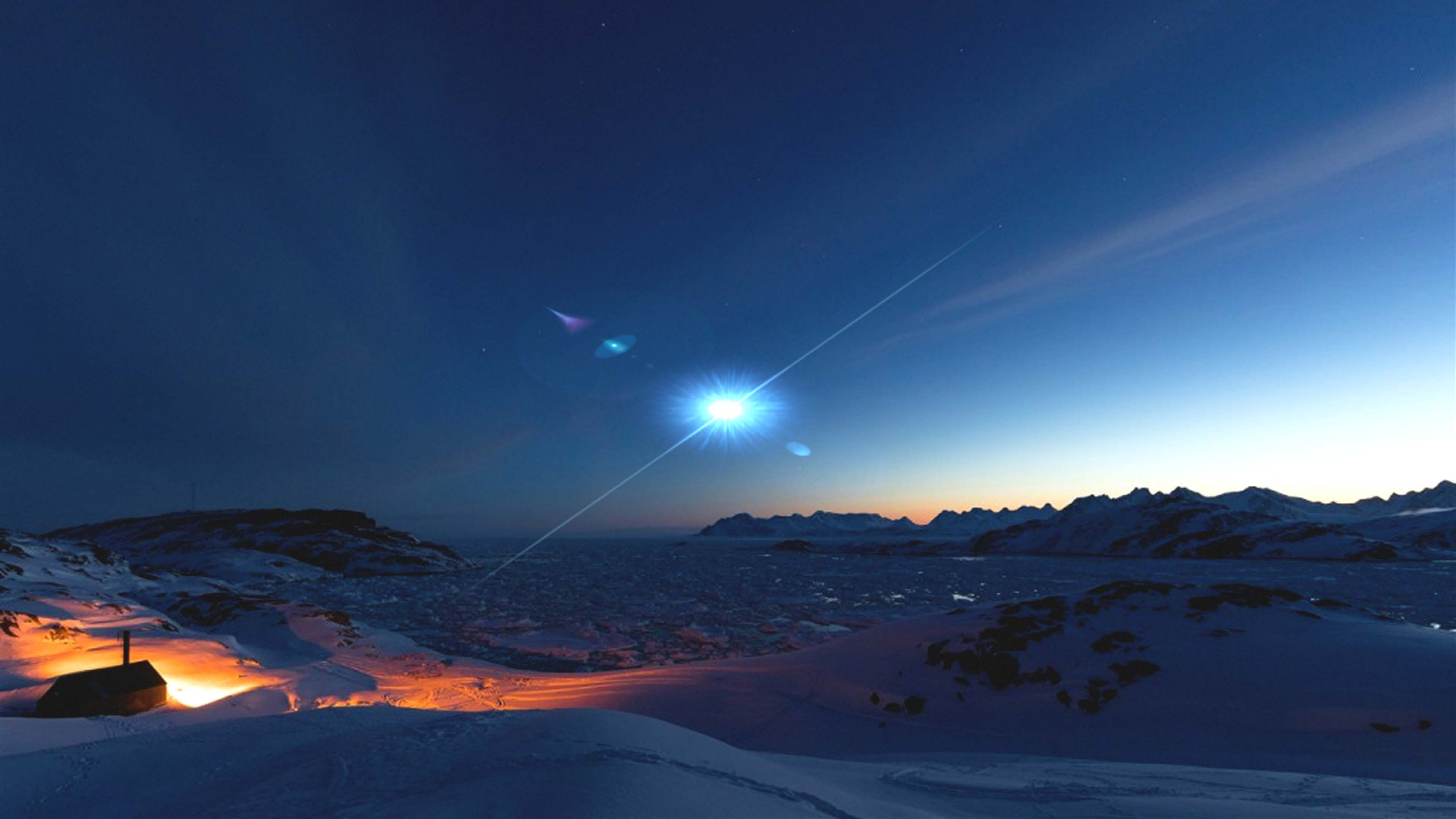}\\ \vspace{-4mm}
		\end{minipage}
	}\hspace{-2mm}
	\subfigure[MFDNet]{
		\begin{minipage}[b]{0.49\textwidth}
   	 	    \includegraphics[width=1\textwidth]{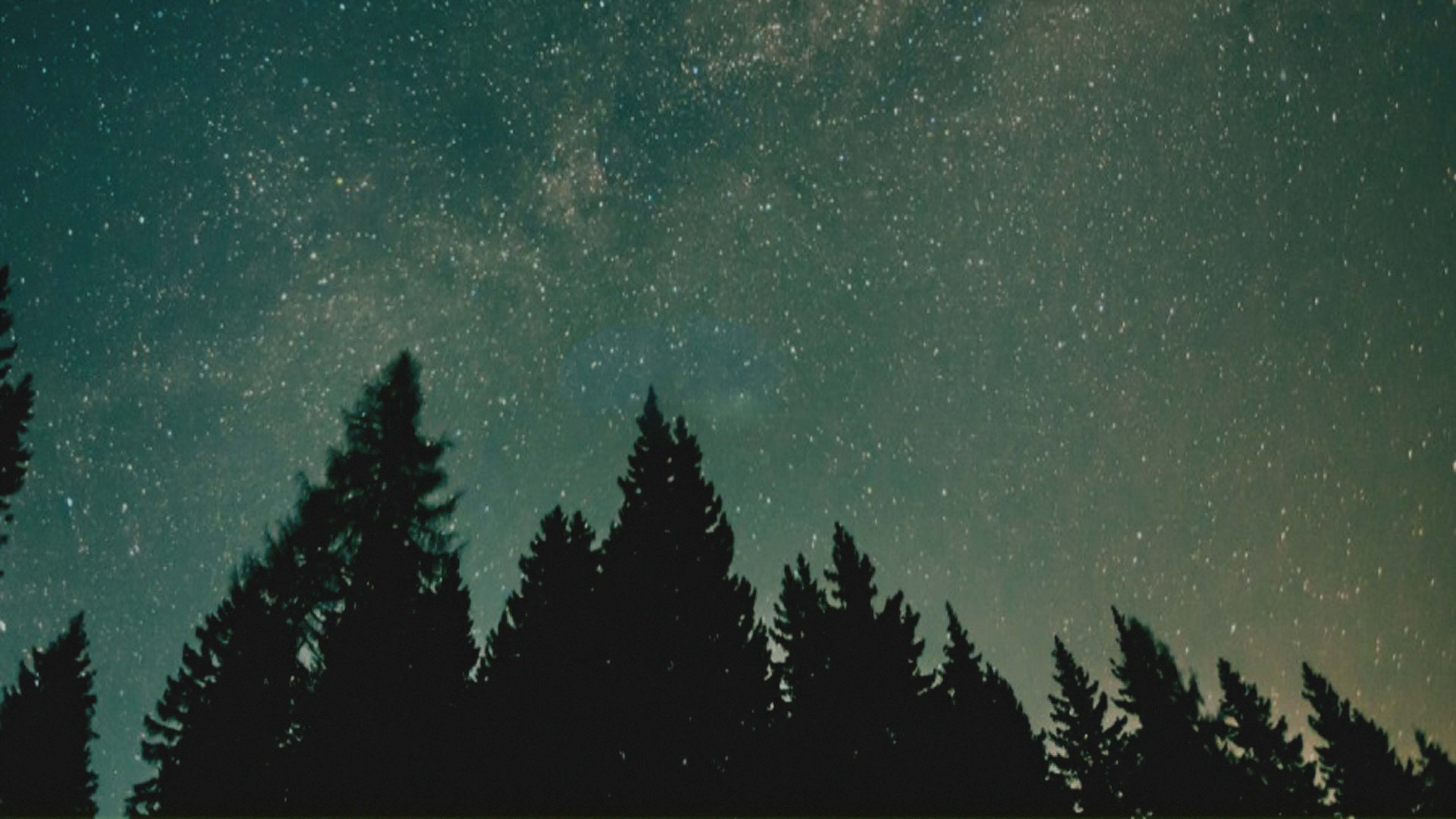}\\ 
   	 	    \includegraphics[width=1\textwidth]{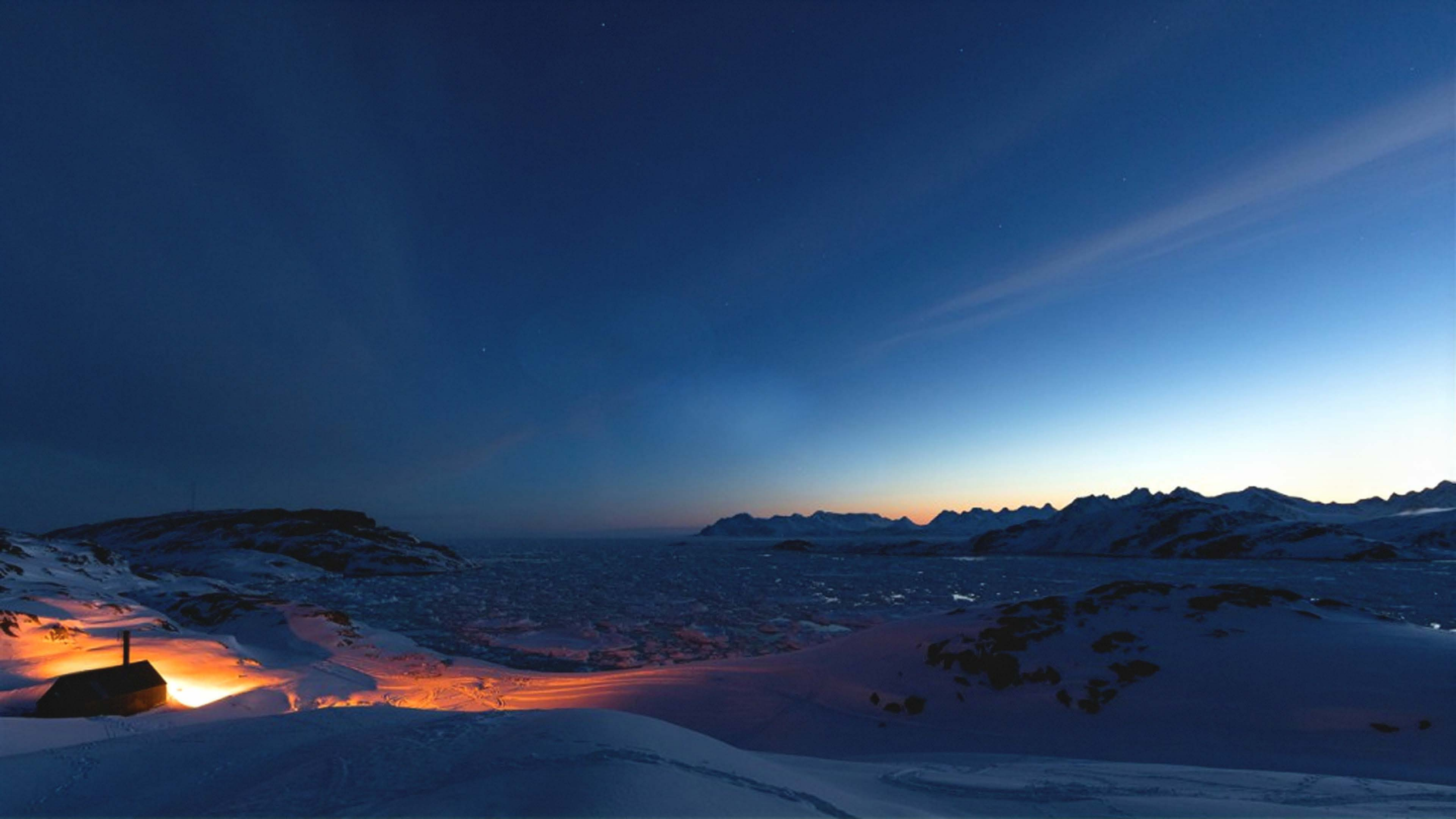}\\ \vspace{-4mm}
		\end{minipage}
		}\hspace{-2mm}
	\end{minipage}\hspace{-2mm}\vspace{-4mm}
	\caption{Flare removal results of our MFDNet on 4K nighttime flare-corrupted images.} \vspace{-4mm}
	\label{fig:4kmore}
\end{figure*}

\subsubsection{Quantitative Evaluation}
Here, we perform a thorough quantitative evaluation of our MFDNet. Following the benchmark provided by \cite{dai2022flare7k}, we use full-reference metrics PSNR, SSIM \cite{wang2004image}, and LPIPS \cite{zhang2018unreasonable} to measure the performance of different methods. In order to better compare the comprehensiveness of the algorithm for removing various flares, we re-evaluate Uformer to make it capable of removing both reflected flares and scattered flares, to replace the original method in \cite{dai2022flare7k}, which can only remove scattered flares. The results are reported in Table \ref{table:overall}. Our MFDNet achieves the highest average PSNR and
SSIM than the previous state-of-the-art methods for real-world nighttime flare removal. For flare removal in synthetic nighttime flare-corrupted images, Our MFDNet performs better in terms of PSNR, SSIM, and LPISP
than the previous state-of-the-art methods. Specifically, our MFDNet achieves 30.79 dB on PSNR, which is at least 0.66 dB higher than all other methods.

\subsubsection{Efficiency Analysis}
In Table \ref{table:efficiency}, we compare the GMACs, parameters, and inference time of different methods on images from $512\times512$ resolution to 4K resolution. Each result in Table \ref{table:efficiency} is an average of 100 tests on an NVIDIA GTX 2080Ti GPU with 11G RAM, where the N.A. denotes that the method cannot handle the input image of this size, and OOM means that the method causes the out-of-memory issue for this specific resolution. As shown in Figure \ref{fig:teaser} and Table \ref{table:efficiency}, our proposed MFDNet outperforms other methods by a significant margin in terms of inference time and GMACs, while also achieving superior deflare performance than others. Specifically, for $512\times512$ resolution images, the GMACs of our method are 18.25G, which is much smaller than the 57.61G of the second-best Restormer \cite{zamir2022restormer}. Our inference time is 0.034s, which is much smaller than the 0.052s of the second-best Dai \cite{dai2022flare7k}. Furthermore, our MFDNet can process images of any size from $512\times512$ to 4K resolution. MPRNet \cite{zamir2021multi} and Uformer \cite{wang2022uformer} can only handle $512\times512$ and $1024\times1024$ resolution images, and none of those state-of-the-art methods can handle 4K resolution images.

Currently, there are no high-resolution nighttime flare image datasets, and the Flare7K dataset that we used for training and testing is at $512\times512$ resolution. To evaluate our MFDNet's performance on 4K resolution images, we followed the pipeline of \cite{dai2022flare7k} by combining flare artifacts into real 4K nighttime images to generate 4K flare-corrupted images. The results of our method on 4K images are illustrated in Figure \ref{fig:4kmore}. Despite the use of $512\times512$ resolution images for our training, our MFDNet can still remove flare on 4K images. This demonstrates the effectiveness and efficiency of our method in removing flare by decoupling the illumination information and content information of the image.

\begin{figure*}[!htb]
	\centering
	\begin{minipage}[b]{1\textwidth}
	\subfigure[Input]{
		\begin{minipage}[b]{0.19\textwidth}
			\includegraphics[width=1\textwidth]{real_more/input/input_000019.png}\\ \vspace{-4mm}
		\end{minipage}
	}\hspace{-2mm}
	\subfigure[w/o FETB+FRTB]{
		\begin{minipage}[b]{0.19\textwidth}
		    \includegraphics[width=1\textwidth]{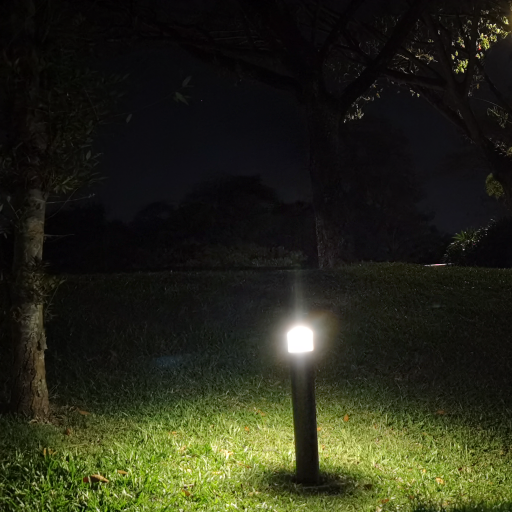}\\ \vspace{-4mm}
		\end{minipage}
	}\hspace{-2mm}
	\subfigure[w/o FECB+FDCB]{
		\begin{minipage}[b]{0.19\textwidth}
		    \includegraphics[width=1\textwidth]{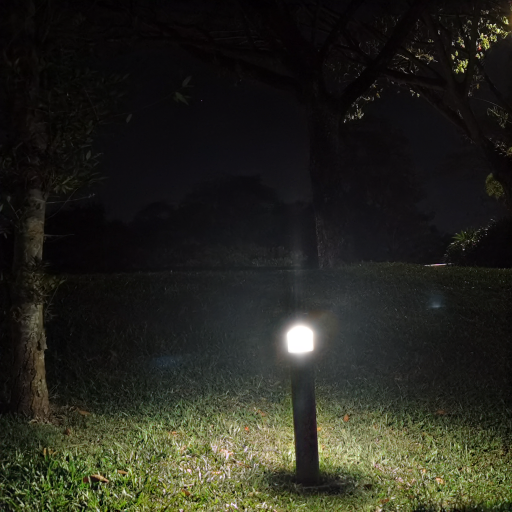}\\\vspace{-4mm}
		\end{minipage}
	}\hspace{-2mm}
	\subfigure[w/o FAB]{
		\begin{minipage}[b]{0.19\textwidth}
		    \includegraphics[width=1\textwidth]{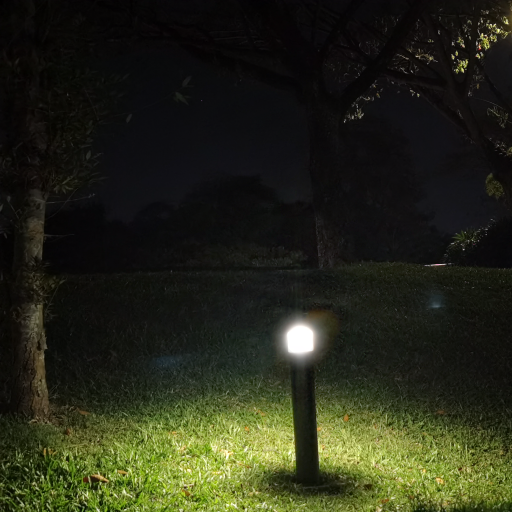}\\ \vspace{-4mm}
		\end{minipage}
	}\hspace{-2mm}
	\subfigure[MFDNet]{
		\begin{minipage}[b]{0.19\textwidth}
		    \includegraphics[width=1\textwidth]{real_more/ours/00019_blend.png}\\ \vspace{-4mm}
		\end{minipage}
	}\hspace{-2mm}
	\end{minipage}\vspace{-4mm}
	\caption{Visual results for the ablation studies.} \vspace{-4mm}
	\label{fig:ab}
\end{figure*}

\begin{table}[h]
\begin{center}
\caption{Ablation experiments of individual components in our MFDNet on Flare7K dataset.}
\adjustbox{width=\columnwidth}{
\begin{tabular}{c c| c c c c}
\toprule[0.15em]
\multicolumn{2}{c|}{Dataset$\backslash$Network} & w/o FETB+FRTB & w/o FECB+FDCB   & w/o FAB & MFDNet \\
\midrule[0.15em]
~ & PSNR~$\uparrow$ &  26.36 &  24.94  & 26.70  & \textbf{26.98} \\
Real-world  & SSIM~$\uparrow$&  0.890  &  0.872  &  0.894 & \textbf{0.895}\\
  & LPIPS~$\downarrow$   &  0.054  &  0.066 &  0.053  &  \textbf{0.051}\\
\midrule[0.1em]
~ & PSNR~$\uparrow$  &  30.32  &  25.95  & 30.61  & \textbf{30.79} \\
Synthetic  & SSIM~$\uparrow$ &  0.963  &  0.925  &  0.964 & \textbf{0.966} \\
  & LPIPS~$\downarrow$  &   0.021 &  0.046 &  0.021  & \textbf{0.019}  \\
\bottomrule[0.15em]
\end{tabular}
}\vspace{-4mm}
\label{table:ablation}
\end{center}
\end{table}\vspace{-4mm}

\subsection{Ablation Studies}
We conduct ablation studies to evaluate the contributions of the different components of our proposed method. Specifically, we measure the contributions of the Feature Extraction Transformer Block (FETB) and the Feature Refinement Transformer Block (FRTB), the Feature Encoder Convolution Block (FECB) and the Feature Decoder Convolution Block (FDCB), and the Feature Aggregation Block (FAB). The results from Table \ref{table:ablation} and Figure \ref{fig:ab} show that these components of our MFDNet are all effective for flare removal. 

The Transformer block for global feature extraction and refinement can improve the deflare performance, because global information is crucial in accurately identifying the flare. Additionally, the convolution-based encoder-decoder structure provides substantial capabilities to remove flare artifacts and restore more image details. As can be seen from Table \ref{table:ablation} and Figure \ref{fig:ab}, after removing the FECB and FEDB, the deflare effect of the model drops significantly, indicating that in the task of flare removal, it is very important to capture the detailed information of the input flare-corrupted image. Moreover, the FAB helps to integrate information from different frequency bands to reconstruct the final flare-free image.

\begin{figure}[t]
\centerline{\includegraphics[width=1 \linewidth]{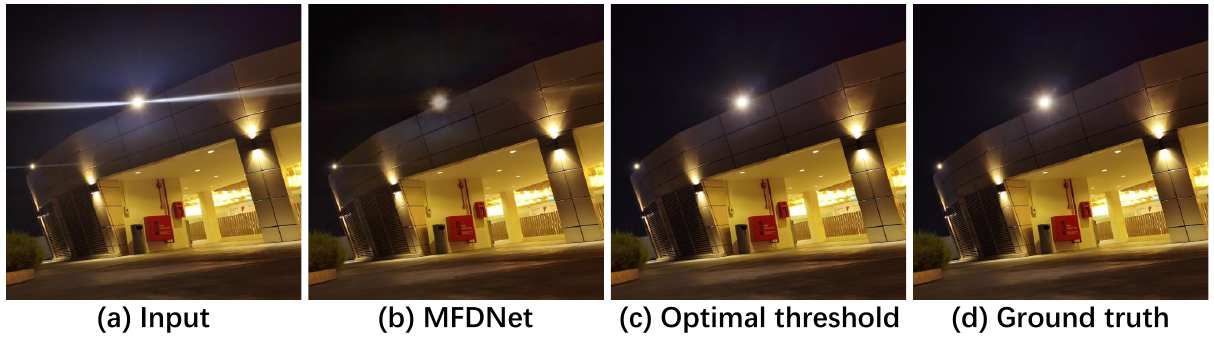}}

\caption{Visual results for the limitation case. From left to right, the sequence includes the input flare image, the result of our MFDNet, our enhanced result after applying the optimal luminance threshold, and the ground truth image.}
\vspace{-4mm}
\label{fig:limit}

\end{figure}

\subsection{Limitations}
However, the proposed method exhibits some limitations that need to be addressed. Firstly, if the light source is diminutive and faint, accompanied by glare flare, the contrast between the light source and adjacent flare areas may not be distinctly pronounced. Under such circumstances, our approach might not fully restore the light source. To address this issue, we might manually determine an optimal luminance threshold for saturation during restoration as shown in Figure \ref{fig:limit}. In future work, we will endeavor to resolve this challenge by employing adaptive techniques to recover the light source. Secondly, our method may fail to completely remove flare shimmers with very high brightness and extensive shapes, probably because they have obvious texture features in the high-frequency band. We intend to resolve these shortcomings in future studies.

\section{Conclusion} \label{conclusion}

We propose a highly efficient Multi-Frequency Deflare
Network (MFDNet) for the nighttime flare removal task, which significantly reduces the computational burden when handling high-resolution images while simultaneously removing flare artifacts successfully. We disentangle illumination information from content information by decomposing the input image with the Laplacian Pyramid. We leverage Transformer's ability to capture long-range pixel dependencies and convolutional neural networks' capability to capture local features to remove flare comprehensively. We developed a hierarchical fusion reconstruction step to adaptively refine high-frequency components, to generate a final flare-free image. Dividing frequency bands to process flare-corrupted images can substantially decrease computational complexity. We demonstrate through extensive experiments that our method is superior and more efficient than other state-of-the-art nighttime flare removal methods.

\bibliographystyle{spmpsci}
\bibliography{ref} 


\end{document}